\documentclass[10pt,twocolumn,letterpaper]{article}
\usepackage{lineno}
\usepackage[pagenumbers]{cvpr} % To force page numbers, e.g. for an arXiv version

\DeclareMathOperator*{\argmin}{arg\,min}

\definecolor{cvprblue}{rgb}{0.21,0.49,0.74}
\usepackage[pagebackref,breaklinks,colorlinks,allcolors=cvprblue]{hyperref}

\def\paperID{5580} % *** Enter the Paper ID here
\def\confName{CVPR}
\def\confYear{2025}

\usepackage{booktabs}       % 更好的表格线
\usepackage{siunitx}        % 数字对齐
\usepackage{multirow}       % 合并行
\usepackage{amsmath}        % 公式居中
\usepackage{array} % 在导言区引入 array 包
\usepackage{graphicx} % 在导言区引入 graphicx 包
\usepackage{amssymb}        % 用于打勾
\usepackage{bbding}
\usepackage{pifont}
\usepackage{wasysym}
\usepackage{utfsym}
\usepackage{fontawesome}
\usepackage{colortbl}
\usepackage{tikz}
\usepackage{balance}
\usepackage{flushend}
\definecolor{red}{HTML}{D2042D}
\newcommand*{\circled}[1]{\lower.7ex\hbox{\tikz\draw (0pt, 0pt)%
    circle (.38em) node {\makebox[1em][c]{\small #1}};}}

\title{Don't Shake the Wheel: Momentum-Aware Planning in\\ End-to-End Autonomous Driving}

\author{Ziying Song$^{1,2,\dagger}$, Caiyan Jia$^{1,2,\star}$, Lin Liu $^{1,2}$, Hongyu Pan$^{3}$, Yongchang Zhang$^{3}$,\\ Junming Wang$^{3,7}$, Xingyu Zhang$^{3}$, Shaoqing Xu$^{4}$, Lei Yang$^{5}$, Yadan Luo$^{6,\star}$\vspace{1ex}\\
$^1$School of Computer Science and Technology, Beijing Jiaotong University\\
$^2$Beijing Key Laboratory of Traffic Data Mining and Embodied Intelligence\\
$^3$Horizon Robotics
$^4$University of Macau
$^5$THU
$^6$The University of Queensland
$^7$HKU
\\
\tt\small\{songziying, cyjia\}@bjtu.edu.cn, y.luo@uq.edu.au.
\vspace{-1ex}
}

\begin{document}
\maketitle
\let\thefootnote\relax\footnotetext{$^\dagger$ Intern of Horizon Robotics, $^{\star}$ Corresponding author.}
\begin{abstract}

% \blfootnote{$\dagger$ Intern of Horizon Robotics. $$ Corresponding Author.}
\noindent
End-to-end autonomous driving frameworks enable seamless integration of perception and planning but often rely on one-shot trajectory prediction, which may lead to unstable control and vulnerability to occlusions in single-frame perception. To address this, we propose the Momentum-Aware Driving (MomAD) framework, which introduces trajectory momentum and perception momentum to stabilize and refine trajectory predictions. MomAD comprises two core components: (1) Topological Trajectory Matching (TTM) employs Hausdorff Distance to select the optimal planning query that aligns with prior paths to ensure coherence; (2) Momentum Planning Interactor (MPI) cross-attends the selected planning query with historical queries to expand static and dynamic perception files. This enriched query, in turn, helps regenerate long-horizon trajectory and reduce collision risks. To mitigate noise arising from dynamic environments and detection errors, we introduce robust instance denoising during training, enabling the planning model to focus on critical signals and improve its robustness. We also propose a novel Trajectory Prediction Consistency (TPC) metric to quantitatively assess planning stability. Experiments on the nuScenes dataset demonstrate that MomAD achieves superior long-term consistency ($\geq 3s$) compared to SOTA methods. Moreover, evaluations on the curated Turning-nuScenes shows that MomAD reduces the collision rate by 26\% and improves TPC by 0.97m (33.45\%) over a 6s prediction horizon, while closed-loop on Bench2Drive demonstrates an up to 16.3\% improvement in success rate.
% These results highlight MomAD’s enhanced stability, robustness, and responsiveness in dynamic driving conditions. 
The source code is available at 
% \href{https://github.com/adept-thu/MomAD}{github}.
 \url{https://github.com/adept-thu/MomAD}.
\end{abstract}

\begin{figure}[t]
\centering
 \includegraphics[width=1.0\linewidth]{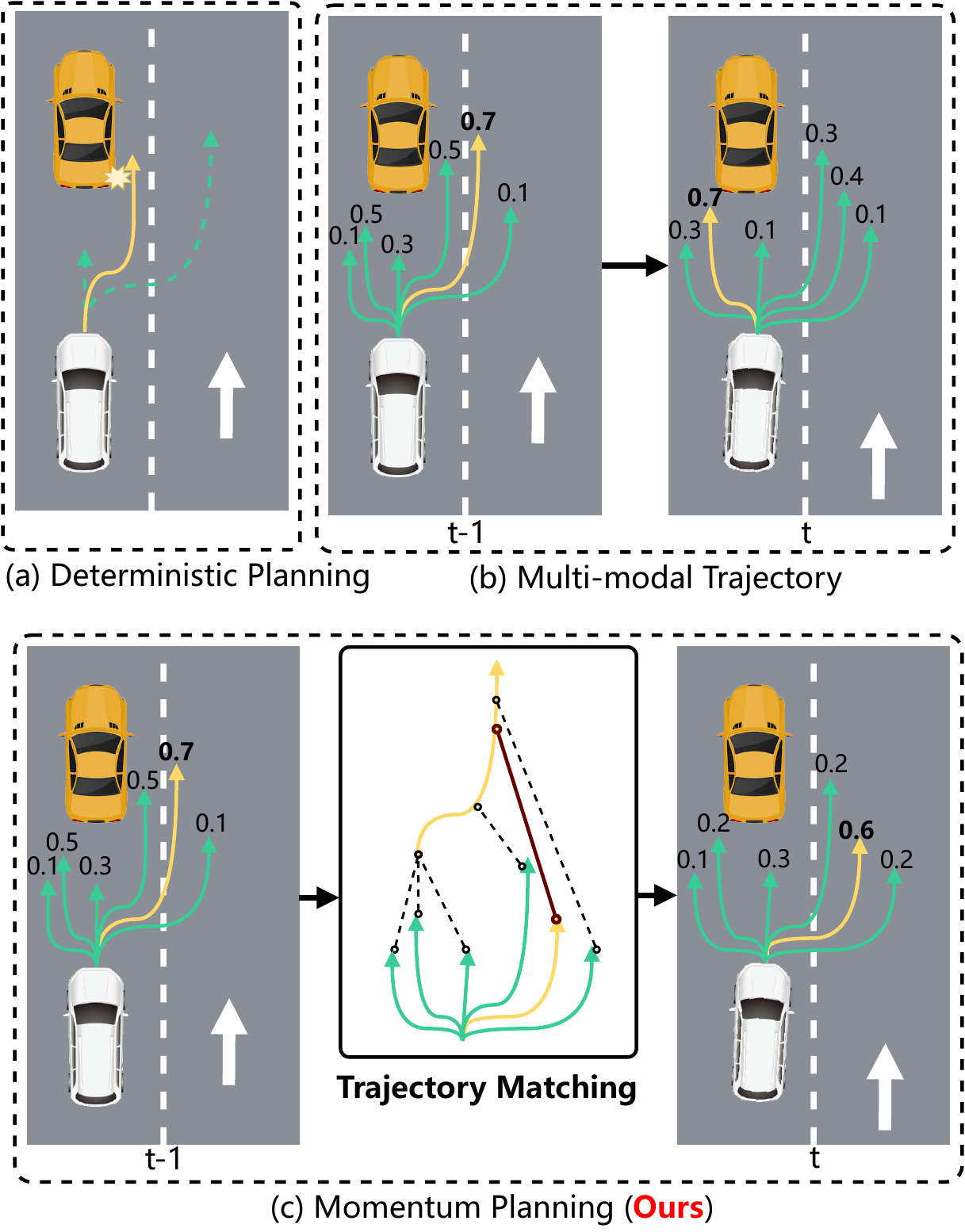}\vspace{-1ex}
\caption[ ]{\textbf{(a)}  \textbf{Deterministic Planning} %Some methods
~\cite{uniad, jiang2023vad, fusionad, ST_P3} predicts deterministic trajectories, but lacks action diversity, posing safety risks. \textbf{(b)} \textbf{Multi-modal Trajectory Planning} 
% SOTA methods~\cite{sun2024sparsedrive,chen2024vadv2,hedrive_zhangxingyu} select the highest-scoring trajectory from multi-modal trajectories, but they inadequately address stability and consistency, potentially resulting in vehicle trembling and compromised safety during driving.
%The methods
~\cite{sun2024sparsedrive,chen2024vadv2,hedrive_zhangxingyu} selects the highest-scoring trajectory %from 
among the multi-modal trajectories, yet fails to ensure stability and consistency, having risks in vehicle trembling. % and compromising safety.
\textbf{(c)} \textbf{Momentum Planning} %We propose MomAD framework, which 
leverages the trajectory and perception momentum
to enhance current planning through historical guidance to overcome temporal inconsistency.

}
\label{fig:motivation}
\end{figure}

\section{Introduction}
\label{introduction}
% benefits of en2end
Autonomous driving~\cite{song2024robustness,wang2023multi,LiHongyange2etpamisurvey} has undergone a transformative shift from modular, manually crafted pipelines to a more integrated, end-to-end paradigm~\cite{uniad,jiang2023vad,MultiFusion}. Unlike traditional approaches that handle tasks like detection, tracking, mapping, motion prediction, and planning in isolation, the end-to-end framework emphasizes seamless integration. By prioritizing planning, it strategically directs information from upstream perception modules, thereby enhancing robustness and reliability in dynamic driving environments.

% existing end2end
Achieving high-quality planning in end-to-end frameworks hinges on accurately predicting the future trajectory prediction for the ego vehicle~\cite{uniad,jiang2023vad,chen2024vadv2,sun2024sparsedrive,reinforcement_motion_planning_survey,MotionPlanning_TITS_survey}. Such future prediction requires a long-horizon understanding of both static and dynamic environmental factors, including map elements and interactions with surrounding agents. For instance, UniAD \cite{uniad} queries the ego context from detailed bird’s-eye-view (BEV) maps at each timestamp, while VAD \cite{jiang2023vad} uses an ego query to retrieve surrounding context. The retrieved information then informs the planner, which predicts a deterministic trajectory for the vehicle, as illustrated in Figure~\ref{fig:motivation} (a). Nevertheless, optimal trajectory prediction is inherently \textit{stochastic} due to the unpredictability of other road users’ intentions, varying road conditions, and the ambiguity introduced by human driving behaviors.  This stochastic nature complicates the regression target, making deterministic predictions suboptimal and even risk-prone, potentially leading to severe collisions. To mitigate these uncertainties, methods such as VADv2~\cite{chen2024vadv2} and SparseDrive~\cite{sun2024sparsedrive} leverage probabilistic modeling to capture the continuous planning action space, producing multi-modal trajectories that consider various possible behaviors of road agents, as shown in Figure \ref{fig:motivation} (b). While effective, these multi-modal approaches are typically \textit{one-shot} and solely on the current perception frame. This limitation makes them susceptible to occlusion or loss of key visual cues, which can degrade multi-modal trajectory quality. Additionally, without temporal consistency, consecutive trajectories may lack coherence, causing unstable vehicle control and introducing undesirable directional shifts and oscillations. 

% ---- old version --- 

% Planning is one of the most crucial modules and tasks in end-to-end autonomous driving, responsible for generating the ego vehicle's future trajectory based on perception and prediction~\cite{LiHongyange2etpamisurvey}. Some methods ~\cite{uniad, jiang2023vad, fusionad, ST_P3} directly regress actions to predict a deterministic trajectory, as shown in Figure ~\ref{fig:motivation} (a). However, they often lack diversity, potentially resulting in intermediate actions that pose safety risks in complex driving environments ~\cite{chen2024vadv2}.  
% As shown in Figure ~\ref{fig:motivation} (b), SOTA (state-of-the-art) methods \cite{sun2024sparsedrive, chen2024vadv2, hedrive_zhangxingyu} use multi-modal trajectory planning, which is a mainstream approach currently, to obtain diverse paths. Although they select the highest-scoring trajectory from the multi-modal proposals as the final plan and have achieved SOTA performance, their trajectory predictions may exhibit oscillations, leading to temporal inconsistency. This invokes the question of how to solve \textbf{temporal inconsistency} caused by multi-modal trajectory planning? 

To stabilize trajectory prediction, we draw inspiration from human driving behaviors and introduce the concept of momentum into autonomous driving.  In physics, momentum reflects an object's tendency to maintain its velocity based on speed and direction \cite{momentum}. Analogously, in driving, momentum captures the smooth, forward progression of movement informed by past trajectories and modulated by present conditions. As illustrated in Figure \ref{fig:motivation} (c), by explicitly integrating historical trajectories with current predictions, we aim to achieve smoother and more coherent planning outcomes. To this end, we propose an end-to-end Momentum-Aware Driving (MomAD) framework, which incorporates momentum awareness to deliver stable and responsive planning in driving scenarios. MomAD interprets momentum on two levels: (1) \textit{trajectory momentum}: By aligning candidate multi-modal trajectory with prior predictions, abrupt shifts in ego vehicle’s path can be minimized, ensuring consistent control and a more comfortable driving experience. (2) \textit{perception momentum}: By aggregating historical context and attending to map elements and surroundings over time, the model broadens its perspective, capturing subtle agent intentions missed in single-frame observations. To implement these ideas, we introduce (1) \textbf{Topological Trajectory Matching (TTM)}: we first use this module to minimize planning discrepancies across time steps by employing the Hausdorff Distance to identify multi-modal trajectory proposals that best align with past planning results. This approach ensures temporal coherence by preventing excessive deviation from previous trajectories. (2)\textbf{ Momentum Planning Interactor (MPI)}: Since the selected trajectory may still be biased toward the current perception and sacrifice long-horizon considerations, we cross-attends the current best planning query with historical plan queries, which implicitly convey critical long-term ego-temporal, ego-agent, and ego-map information as key and value vectors. This interaction enriches the current query with long-horizon perception momentum, improving its context awareness. To enhance robustness against environmental noise and perception errors, we incorporate a Robust Instance Denoising Module in the perception stage. By introducing controlled perturbations during training, the model learns to denoise perception inputs, achieving resilience to dynamic changes and misdetections.

To evaluate the planning stability of MomAD, we propose a new Trajectory Prediction Consistency (TPC) metric to measure consistency between predicted and historical trajectories. Experiments demonstrate that MomAD can maintain long-term consistency ($\geq$ 3 seconds). Given that most scenes in nuScenes involve straight roads, which limit the assessment of temporal inconsistency, we curated a Turning-nuScenes validation set from turning scenarios within the nuScenes dataset to provide a more challenging evaluation, where our approach outperforms state-of-the-art end-to-end frameworks. For example, our MomAD reduces the collision rate by 26\% and the TPC by 0.97m compared to SparseDrive~\cite{sun2024sparsedrive} for a 6-second horizon prediction in the Turning-nuScenes validation set.

\begin{figure*}[t]
\centering
 \includegraphics[width=1.0\linewidth]{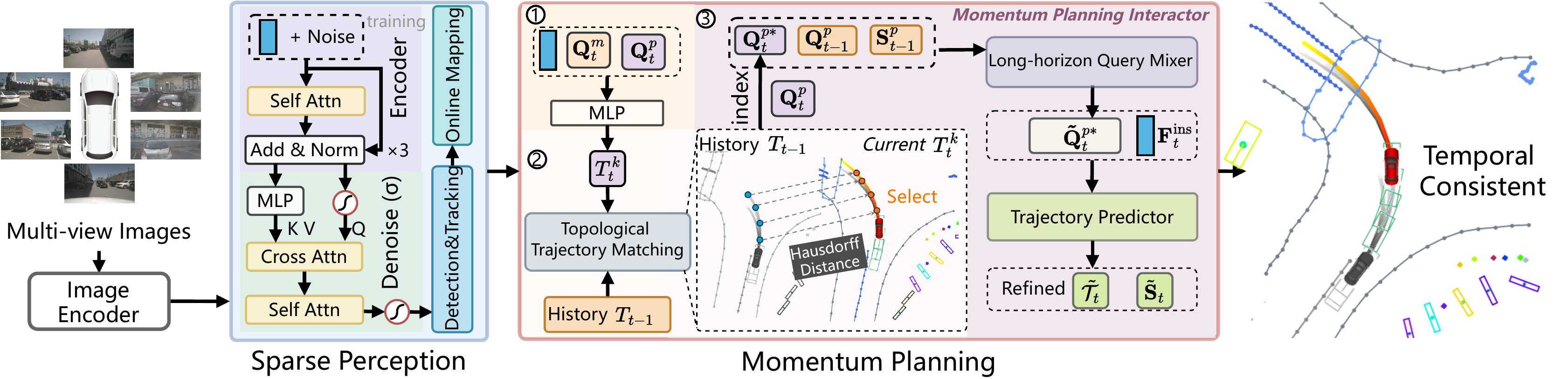}
\caption[ ]{
\textbf{The overall architecture of MomAD.} MomAD, as a multi-modal trajectory end-to-end autonomous driving method, first encodes multi-view images into feature maps, then learns a sparse scene representation through a robust instance denoising via perturbation 
module, and finally performs a momentum planning through Topological Trajectory Matching (TTM) module and Momentum Planning Interactor (MPI) module to accomplish planning tasks. Our approach addresses critical challenges of stability and robustness in  dynamic driving conditions.% of end-to-end autonomous driving systems.
%Our proposed 

}
\label{fig:main}
\end{figure*}

\begin{figure}[t]
\centering
 \includegraphics[width=1.0\linewidth]{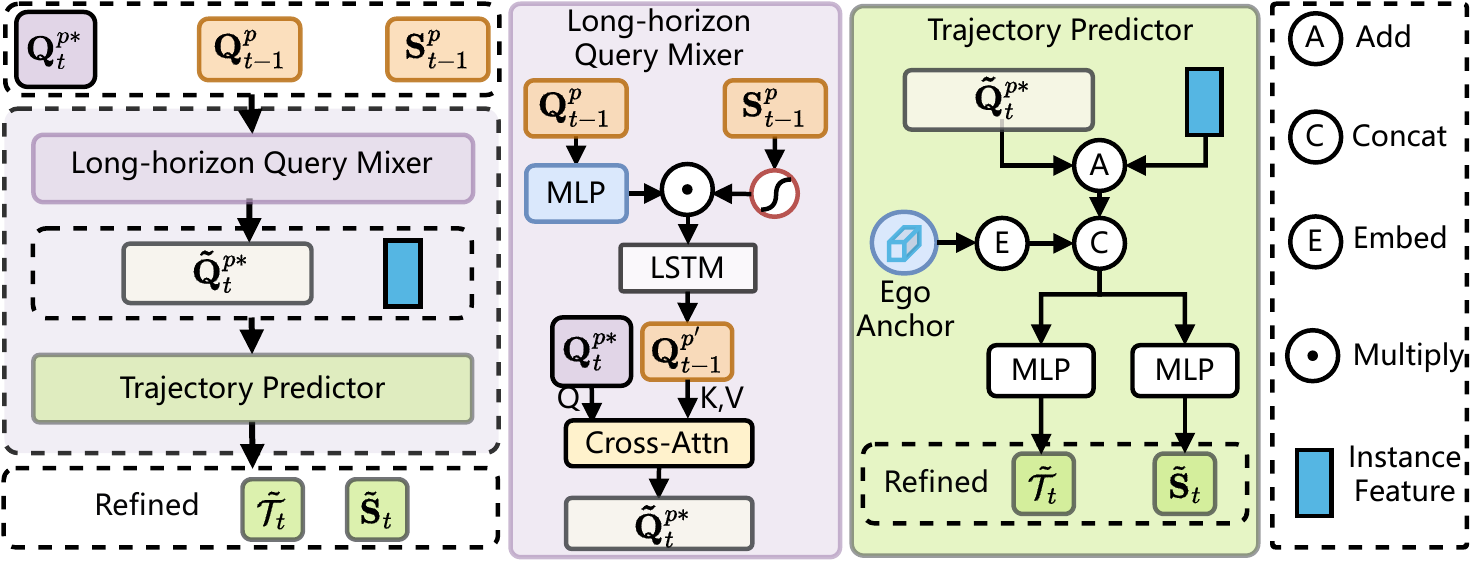}
\caption[ ]{
% \textbf{The architecture of Momentum Planning Interactor.} After accomplishing TTM, 
% a selected planning query $\mathbf{Q}_{t}^{p*}$ is obtained via %which is the most consistent with 
% the historical multi-modal trajectory queries $\mathbf{Q}_{t-1}^p$. 
% Then, the historical trajectory scores $\mathbf{S}_{t-1}^p$ and $\mathbf{Q}_{t-1}^p$ are inputted into the Long-horizon Query Mixer, passing through %where the weighted historical trajectory query $\mathbf{Q}_{t-1}^p$ using the scores is fed into 
% an LSTM layer to simulate temporal evolution $\mathbf{Q}_{t-1}^{p'}$. $\mathbf{Q}_{t}^{p*}$ serves as the Query to selectively query $\mathbf{Q}_{t-1}^{p'}$ for information fusion. Finally,  $\mathbf{\tilde{Q}}_{t}^{p*}$ along with the instance feature $\mathbf{F}_t^{\operatorname{ins}}$ are sent to Trajectory Predictor for trajectory predictions.
% 
% }
\textbf{The illustration of Momentum Planning Interactor (MPI).} MPI cross-attends a selected planning query with historical queries to expand static and dynamic perception files, resulting in an enriched query that improves long-horizon trajectory generation and reduces collision risks.
}
\label{fig:sub2}
\end{figure}

\section{Related Work}
\label{related_work}
End-to-end autonomous driving, which learns directly from raw sensor data to generate planning trajectories or driving commands, eliminates manual feature extraction \cite{song2024robustness,LiHongyange2etpamisurvey}. 
% It enables more efficient data utilization and greater adaptability to diverse driving scenarios. 
% Early works ~\cite{e_2_e_TCP,e_2_e_Codevilla_2019_ICCV,e_2_e_ICRA2018,e_2_e_Zhang_2021_ICCV_Coach,codevilla2018end,IN-Planner} excluded intermediate tasks, such as perception and motion prediction, which were characterized by limited interpretability and difficulties in optimization.  
% In addition, some methods \cite{xu2024m2da,TransFuser,interfuser,thiktwice,gpt_driver,jia2023driveadapter,shao2023reasonnet,chitta2021neat} has focused on closed-loop end-to-end driving in simulators \cite{CARLA}. However, a domain gap persists between simulated environments and the real world, particularly in sensor data and agent motion states \cite{bevplanner}. 
End-to-end autonomous driving methods \cite{e_2_e_TCP,e_2_e_Codevilla_2019_ICCV,e_2_e_ICRA2018,e_2_e_Zhang_2021_ICCV_Coach,codevilla2018end,IN-Planner,mao2023gptdriver,hedrive_zhangxingyu,uniad,jiang2023vad,zheng2024genad,fusionad,zhang2024graphad,doll2024dualad,chen2025ppad,guo2024uad,yang2024deepinteraction++,Senna,PIP,su2024difsd,cheng2024rethinking,li2024hydra,FF} have garnered increased attention. UniAD~\cite{uniad} effectively integrates information from various preceding tasks, including perception, prediction, and planning modules to assist in trajectory planning, and achieves significant performance improvements. VAD~\cite{jiang2023vad} models driving scenarios as fully vectorized representations and employs explicit instance-level planning constraints to enhance planning safety.  
% DualAD \cite{doll2024dualad} disentangles dynamic agents and static elements, compensating for motion effects and flexibly propagating the belief state over time.
% PPAD~\cite{chen2025ppad} considers the timestep-wise interaction to better integrate prediction and planning, with the final output being a deterministic trajectory.
However, they \cite{uniad,jiang2023vad,ST_P3, chen2025ppad, doll2024dualad} adopt a deterministic approach to trajectory prediction, which fails to account for trajectory diversity and may introduce risks due to intermediate regression results. VADv2 \cite{chen2024vadv2} proposes probabilistic planning to address the limitations of deterministic trajectory prediction, enabling multi-modal trajectory predictions.
% Hydra-MDP \cite{li2024hydra} propose a universal framework of end-to-end multi-modal planning via multi-target hydra-distillation, allowing the model to learn from both rule-based planners and human drivers in a scalable manner.
%Furthermore, SparseDrive~\cite{sun2024sparsedrive}
Building upon the multi-modal trajectory planning framework, SparseDrive~\cite{sun2024sparsedrive} designs planning and motion prediction modules to achieve SOTA performance and efficiency on the nuScenes dataset.
However, while multi-modal trajectory planning methods \cite{chen2024vadv2,sun2024sparsedrive,hedrive_zhangxingyu} have achieved SOTA performance, they overlook the temporal inconsistency caused by maximum score offsets. Regarding temporal inconsistency, existing methods only address temporal instance characteristics up to now, and they entirely overlook the issue of temporal planning consistency. In this work, we focus on this issue, aiming to address it using the concept of momentum planning.

\section{Method}
\label{method}
\textbf{Framework Overview.} Figure \ref{fig:main} presents an overview of the proposed MomAD system, which integrates sparse perception and momentum-aware planning. To capture key dynamic and static instances interacting with the ego vehicle, the sparse perception module builds upon the SparseDrive \cite{sun2024sparsedrive} to encode multi-view image features, which are aggregated into instance features $\mathbf{F}_t^{\operatorname{ins}}$ for road agents and map elements at time step $t$. These features are obtained by sampling keypoints around the anchor boxes and polylines, feeding into the detection/tracking and online mapping blocks for accurate predictions. The core of MomAD is the joint motion and momentum-aware planning module, which comprises two main components: (1) Topological Trajectory Matching (Sec \ref{sec:ttm}), which explicitly selects the candidate trajectory that best matches the prior path among all multi-modal trajectories to ensure temporal coherence; and (2) Momentum Planning Interaction (Sec \ref{sec:mpi}), which expands the perceptive field by cross-attending the selected candidate trajectory’s planning query with queries from the previous time step in the long-horizon query mixer. This approach provides a broader view of the surrounding environment and the intentions of other agents. The refined query is then processed by the planning head to generate updated multi-modal trajectories. Since the planning module heavily relies on detection and map instance features, we introduce a robust instance denoising via perturbation module (Sec \ref{sec:perturb}) within the sparse perception component during training. This ensures robustness by reducing sensitivity to noisy perception features, enhancing the stability of trajectory prediction and planning.

\subsection{Topological Trajectory Matching (TTM)}\label{sec:ttm}
The proposed TTM module is inspired by the continuity of human driving behavior, where the optimal trajectories $T^*_t$ are influenced by the historical path $T_{t-1}$ to maintain temporal consistency and stability. Let $\mathcal{T}_t = \{T_t^k\}_{k=1}^K$ denote a set of $K$ multi-modal candidate trajectories generated at time step $t$, where each trajectory $T_t^k = \{(x_{t,i}^k, y_{t,i}^k)\}_{i=1}^{N_t}$ consists of $N_t$ predicted waypoints. Typically, $K$ is set to $6\times 3$ to account for six trajectory proposals for each of three possible commands (left, right, and straight), and $N_t$ is chosen as $6$ or $12$ representing 0.5s interval for a 3- or 6-second prediction horizon.

\noindent\textbf{Trajectory Coordinate Transformation.} Since the historical and current predicted trajectories are generated in the ego vehicle's coordinate system at different moments, it is essential to transform them into a common coordinate system for accurate matching. This transformation from moment $t$ to ${t-1}$ is achieved as follows:
\begin{equation}
    \label{eq:shortencode}
    \begin{aligned}
T^{k}_{t} \leftarrow \mathbf{R}_{t-1}^{-1}(T^{k}_{t} - \mathbf{\Gamma}_{t-1}), ~\forall k\in [K],
    \end{aligned}
\end{equation}
where $\mathbf{R}_{t-1}$ and $\mathbf{\Gamma}_{t-1}$ denote the rotation and translation matrix, respectively.

\noindent\textbf{Trajectory Distance Measurement.} Simple Euclidean distance is inadequate for capturing the global alignment of trajectories, as it only measures pointwise proximity and is highly sensitive to local variations. This limitation becomes especially apparent in complex scenarios such as turns or varying point densities, where close points may not represent the alignment of the entire trajectory path. To address these limitations, TTM employs the Hausdorff distance as a more robust metric for evaluating trajectory alignment. The Hausdorff distance captures both local and global trajectory structures by measuring the maximum deviation between two sets of points, effectively quantifying the worst-case alignment between the candidate and historical trajectories. For each candidate trajectory $T^{k}_{t}$, the Hausdorff distance to the historical trajectory $T_{t-1}$ is computed as,
\begin{equation} 
    \begin{aligned} 
        d_1 = \sup_{p \in T_t^k} \inf_{h \in T_{t-1}} \| p - h \|,& \ d_2 = \sup_{h \in T_{t-1}} \inf_{p \in T_t^k} \| h - p \|, \\
         d_H(T_t^k, T_{t-1}) &= \max(d_1, d_2), 
    \end{aligned} 
\end{equation}
where $p\in T_{t}^k$ and $h\in T_{t-1}$ represent waypoints in the candidate and historical trajectories, respectively. The Hausdorff distance considers the furthest point discrepancies between trajectories in both directions, ensuring that even minor global misalignments are captured. TTM then selects the trajectory $T_{t}^{k^*}$ that minimizes this distance:
\begin{equation}
    k^* = \argmin_{k\in[K]} d_{H}(T_t^k, T_{t-1})
\end{equation}
This selection enforces continuity, aligning with historical driving patterns and providing stable trajectory predictions that are less prone to sudden shifts.

\subsection{Momentum Planning Interactor (MPI)}\label{sec:mpi}
While TTM selects the most consistent trajectory $T_t^{k^*}$ based on historical alignment, $T_t^{k^*}$ is solely based on the current perception $\mathbf{F}_{\operatorname{ins}}$, which may lack a comprehensive view of the environment and be sensitive to occlusions. Therefore, the MPI module, as illustrated in Figure~\ref{fig:sub2}, incorporates a \textit{long-horizon query mixer} to enrich the selected planning query $\mathbf{Q}_{t}^{p*}\in\mathbb{R}^{D_q}$ of $T_t^{k^*}$ with historical planning query $\mathbf{Q}_{t-1}^p\in\mathbb{R}^{K\times D_q}$ and the associated planning scores  $\mathbf{S}_{t-1}^p\in\mathbb{R}^{K}$, implicitly capturing a broader understanding of the surrounding context and other agents' intentions over time. Here, $D_q$ represents the latent dimension of planning queries. This enriched planing query $\mathbf{\tilde{Q}}_{t}^{p*}\in\mathbb{R}^{D_q}$ will be combined with instance features to re-generate an improved trajectory $\tilde{\mathcal{T}}_t\in\mathbb{R}^{K\times N_t\times 2}$.

% After implementing \textbf{Topological Trajectory Matching (TTM)}, we obtain the best trajectory from the multi-modal trajectory proposals. However, the best trajectory is merely the result of the current planning, and it inherently contains misalignment, truly failing to address the temporal inconsistency. Therefore, we further propose \textbf{Momentum Planning Interactor}, which refines the current planning query to generate temporal consistent multi-modal trajectories.

% As shown in Figure ~\ref{fig:sub2}, \textbf{Momentum Planning Interactor} primarily consists of two sub-modules, \textbf{Long-horizon Query Mixer} and \textbf{Trajectory Predictor}. Specifically, in \textbf{Long-horizon Query Mixer}, to emphasize the reliable historical features, the historical planning query $Q_{t-1}^{p}$ and the score features $S_{t-1}^{p}$ are retained for the current frame. The score features $S_{t-1}^{p}$ undergo a sigmoid operation to update the score weights, while $Q_{t-1}^{p}$ is processed as shown in the following formula.
% where \textbf{D}($\,\cdot\,,\,\cdot\,$) denotes the dot product.
% The equation in Eq. \ref{eq:re-weight} generates the history planning query $Q_{t-1}^{p'}$, weighted by the score $\sigma (S_{t-1}^{p})$, which serves as prior knowledge for the current planning. 

\noindent \textbf{Long-horizon Query Mixer.} To achieve a robust perception of temporal momentum, the query mixer allows cross-attention between the selected candidate trajectory’s planning query with multi-modal planning queries from the previous time step. The historical planning queries $\mathbf{Q}_{t-1}^{p}$ and associated scores $\mathbf{S}_{t-1}^{p}$ are combined through element-wise interaction and processed with an LSTM to simulate temporal evolution:
\begin{equation}
    \label{eq:re-weight}
    \begin{aligned}
        \mathbf{Q}_{t-1}^{p'} &= \operatorname{LSTM}\left(\sigma(\mathbf{S}_{t-1}^{p})\circ \operatorname{MLP}(\mathbf{Q}_{t-1}^{p})  \right),
    \end{aligned}
\end{equation}
where $\sigma(\cdot)$ indicates the sigmoid function, $\operatorname{MLP}:\mathbb{R}^{D_q}\mapsto\mathbb{R}^{D_q}$ the linear transform and $\circ$ the element-wise product. The $\operatorname{LSTM}$ processes this interaction, producing a surrogate multi-modal query $\mathbf{Q}_{t-1}^{p'}\in\mathbb{R}^{K \times D_q}$ that captures the temporal evolution of planning queries. To aggregate historical information, the current planning query $\mathbf{Q}_{t}^{p*}$ is used as a query in a cross-attention module. The result $\mathbf{\tilde{Q}}_{t}^{p*}$ incorporates long-term spatiotemporal context, which is further combined with the planning instance features and the encoded ego-anchor position information to inform the subsequent trajectory predictor:
\begin{equation}
    \label{eq:crossattn}
    \begin{aligned}
        \mathbf{\tilde{Q}}_{t}^{p*} &= \operatorname{Attention}(\mathbf{Q}_{t}^{p*}, \mathbf{Q}_{t-1}^{p'}, \mathbf{Q}_{t-1}^{p'}).\\     
        \tilde{\mathcal{T}}_t, \tilde{\mathbf{S}}_t &= \operatorname{PlanHead}(\mathbf{\tilde{Q}}_{t}^{p*}, \mathbf{F}_t^{\operatorname{ins}}),
    \end{aligned}
\end{equation}
The $\operatorname{PlanHead}$ module then generates refined $\tilde{\mathcal{T}}_t, \tilde{\mathbf{S}}_t$. The best trajectory $T^*_t$ is then selected based on the highest scores among multi-modal outputs. Importantly, while the best trajectory is chosen based on the multi-modal trajectory scores, unlike previous selections, the current multi-modal trajectories now fully consider the temporal consistency. This approach provides a stable, temporally-aware planning solution that is robust to occlusions and noise, significantly improving trajectory stability and control in complex driving environments.

\subsection{Robust Instance Denoising via Perturbation}\label{sec:perturb}
Our trajectory prediction and refinement rely heavily on the instance features $\mathbf{F}^{\operatorname{ins}}_t$ of road agents and map elements provided by the sparse perception module. However, due to detector instability and the dynamically changing map, these instance features may be noisy, potentially introducing errors in downstream planning. To enhance the stability of planning against such noisy inputs, we introduce controlled noise perturbations during training and employ a lightweight encoder-decoder transformer block (see Figure \ref{fig:main}) to learn effective denoising. This approach enables the model to distinguish between essential and extraneous features, reducing the impact of perception noise on trajectory predictions. During test-time inference, this denoising capability allows the trajectory predictor to be resilient to fluctuations in instance features. As a result, the model can produce smoother, more stable trajectories even in challenging scenarios with occlusions, temporary obstacles, or misdetections. 
% The integration of instance perturbation-denoising thus reinforces the robustness of our framework, ensuring that the planning module remains consistent and reliable in dynamic driving environments.

% \begin{figure}[htp]
% \centering
%  \includegraphics[width=0.6\linewidth]{images/sub1V3.pdf}
% \caption[ ]{
% The architecture of Encoder and Denoise ($\sigma$) module in Sparse Perception.
% }
% \label{fig:sub1}
% \end{figure}

\begin{table*}[htp]
\centering
  \caption{Planning results on the $\operatorname{nuScenes}$ validation dataset. $^{\dagger}$ denotes evaluation protocol used in $\operatorname{UniAD}$~\cite{uniad}. $^{\ast}$ denotes results reproduced with the official checkpoint. As Ref.~\cite{bevplanner} states, we \textbf{deactivate} the \textbf{ego status} information for a fair comparison.}
  \renewcommand\arraystretch{1.0}
  \setlength{\tabcolsep}{1.4mm} % Adjust column spacing
  \resizebox{\linewidth}{!}{
  \begin{tabular}{lclcccc cccc cccc c}
\toprule
\multirow{2}{*}{$\operatorname{Method}$} & \multirow{2}{*}{$\operatorname{Input}$} & \multirow{2}{*}{$\operatorname{Backbone}$} & \multicolumn{4}{c}{$\operatorname{L2\ (m)}\downarrow$} & \multicolumn{4}{c}{$\operatorname{Col.\ Rate\ (\%)}\downarrow$} & \multicolumn{4}{c}{$\operatorname{TPC\ (m)}\downarrow$} & \multirow{2}{*}{$\operatorname{FPS}\uparrow$} \\
\cmidrule(lr){4-7} \cmidrule(lr){8-11} \cmidrule(lr){12-15}
& & & 1s & 2s & 3s & $\operatorname{Avg.}$ & 1s & 2s & 3s & $\operatorname{Avg.}$ & 1s & 2s & 3s & $\operatorname{Avg.}$ \\
\midrule
% $\operatorname{FF}^{\dagger}$~\cite{FF} & $\operatorname{LiDAR}$ & $\operatorname{PIXOR}$~\cite{yang2018pixor} & 0.55 & 1.20 & 2.54 & \cellcolor{gray!15}1.43 & 0.08 & 0.27 & 1.95 & \cellcolor{gray!15}0.77 & - & - & - & \cellcolor{gray!15}- & - \\
% $\operatorname{EO}^{\dagger}$~\cite{EO} & $\operatorname{LiDAR}$ & $\operatorname{PIXOR}$~\cite{yang2018pixor} & 0.67 & 1.36 & 2.78 & \cellcolor{gray!15}1.60 & 0.04 & 0.09 & 0.88 & \cellcolor{gray!15}0.33 & - & - & - & \cellcolor{gray!15}- & - \\
% $\operatorname{ST\text{-}P3}^{\dagger}$~\cite{ST_P3} & $\operatorname{Camera}$ & $\operatorname{ResNet50}$ & 1.72 & 3.26 & 4.86 & \cellcolor{gray!15}3.28 & 0.44 & 1.08 & 3.01 & \cellcolor{gray!15}1.51 & - & - & - & \cellcolor{gray!15}- & - \\
$\operatorname{UniAD}^{\dagger}$~\cite{uniad} & $\operatorname{Camera}$ & $\operatorname{ResNet101}$ & 0.48 & 0.96 & 1.65 & \cellcolor{gray!15}1.03 & \textbf{0.05} & 0.17 & 0.71 & \cellcolor{gray!15}0.31 & 0.45 & 0.89 & 1.54 & \cellcolor{gray!15}0.96 & 1.8 $\operatorname{(A100)}$ \\
$\operatorname{VAD}^{\dagger}$~\cite{jiang2023vad} & $\operatorname{Camera}$ & $\operatorname{ResNet50}$ & 0.54 & 1.15 & 1.98 & \cellcolor{gray!15}1.22 & 0.10 & 0.24 & 0.96 & \cellcolor{gray!15}0.43 & 0.47 & 0.83 & 1.43 & \cellcolor{gray!15}0.91 & - \\
$\operatorname{SparseDrive}^{\dagger\ast}$~\cite{sun2024sparsedrive} & $\operatorname{Camera}$ & $\operatorname{ResNet50}$ & 0.44 & 0.92 & 1.69 & \cellcolor{gray!15}1.01 & 0.07 & 0.19 & 0.71 & \cellcolor{gray!15}0.32 & 0.39 & 0.77 & 1.41 & \cellcolor{gray!15}0.85 & \textbf{9.0 $\operatorname{(RTX4090)}$} \\
\rowcolor{gray!15} $\operatorname{MomAD\ (Ours)}$$^{\dagger}$ & $\operatorname{Camera}$ & $\operatorname{ResNet50}$ & \textbf{0.43} & \textbf{0.88} & \textbf{1.62} & \textbf{0.98} & 0.06 & \textbf{0.16} & \textbf{0.68} & \textbf{0.30} & \textbf{0.37} & \textbf{0.74} & \textbf{1.30} & \textbf{0.80} & 7.8 $\operatorname{(RTX4090)}$ \\
\midrule
% $\operatorname{ST\text{-}P3}$~\cite{ST_P3} & $\operatorname{Camera}$ & $\operatorname{ResNet50}$ & 1.33 & 2.11 & 2.90 & \cellcolor{gray!15}2.11 & 0.23 & 0.62 & 1.27 & \cellcolor{gray!15}0.71 & - & - & - & \cellcolor{gray!15}- & - \\
$\operatorname{UniAD}$~\cite{uniad} & $\operatorname{Camera}$ & $\operatorname{ResNet101}$ & 0.45 & 0.70 & 1.04 & \cellcolor{gray!15}0.73 & 0.62 & 0.58 & 0.63 & \cellcolor{gray!15}0.61 & 0.41 & 0.68 & 0.97 & \cellcolor{gray!15}0.68 & 1.8 $\operatorname{(A100)}$ \\
$\operatorname{VAD}$~\cite{jiang2023vad} & $\operatorname{Camera}$ & $\operatorname{ResNet50}$ & 0.41 & 0.70 & 1.05 & \cellcolor{gray!15}0.72 & 0.03 & 0.19 & 0.43 & \cellcolor{gray!15}0.21 & 0.36 & 0.66 & 0.91 & \cellcolor{gray!15}0.64 & - \\
$\operatorname{SparseDrive}$~\cite{sun2024sparsedrive} & $\operatorname{Camera}$ & $\operatorname{ResNet50}$ & \textbf{0.29} & 0.58 & 0.96 & \cellcolor{gray!15}0.61 & \textbf{0.01} & \textbf{0.05} & \textbf{0.18} & \cellcolor{gray!15}\textbf{0.08} & \textbf{0.30} & 0.57 & 0.85 & \cellcolor{gray!15}0.57 & \textbf{9.0 $\operatorname{(RTX4090)}$} \\
\rowcolor{gray!15} $\operatorname{MomAD\ (Ours)}$ & $\operatorname{Camera}$ & $\operatorname{ResNet50}$ & 0.31 & \textbf{0.57} & \textbf{0.91} & \cellcolor{gray!15}\textbf{0.60} & \textbf{0.01} & \textbf{0.05} & 0.22 & \cellcolor{gray!15}0.09 & \textbf{0.30} & \textbf{0.53} & \textbf{0.78} & \cellcolor{gray!15}\textbf{0.54} & 7.8 $\operatorname{(RTX4090)}$ \\
\bottomrule
\end{tabular} }
\label{tab_nuscenes_planning}
\end{table*}

\begin{table*}[htp]
\small
\centering
  \caption{Planning results on the \textbf{$\operatorname{Turning\text{-}nuScenes}$} validation dataset. $\operatorname{SparseDrive}$~\cite{sun2024sparsedrive} is a SOTA end-to-end multi-modal trajectory planning method. \textcolor{red}{Red} indicates improvement. We follow the VAD \cite{jiang2023vad} evaluation metric.}\vspace{-2ex}
  \renewcommand\arraystretch{1.0}
  \setlength{\tabcolsep}{1.5mm} % Adjust column spacing
  \resizebox{\linewidth}{!}{
  \begin{tabular}{lcccc cccc cccc}
\toprule
\multirow{2}{*}{$\operatorname{Method}$} & \multicolumn{4}{c}{$\operatorname{L2\ (m)}\downarrow$} & \multicolumn{4}{c}{$\operatorname{Col.\ Rate\ (\%)}\downarrow$} & \multicolumn{4}{c}{$\operatorname{TPC\ (m)}\downarrow$} \\
\cmidrule(lr){2-5} \cmidrule(lr){6-9} \cmidrule(lr){10-13}
& 1s & 2s & 3s & $\operatorname{Avg.}$ & 1s & 2s & 3s & $\operatorname{Avg.}$ & 1s & 2s & 3s & $\operatorname{Avg.}$ \\
\midrule
$\operatorname{SparseDrive}$~\cite{sun2024sparsedrive} & 0.35 & 0.77 & 1.46 & \cellcolor{gray!15}0.86 & 0.04 & 0.17 & 0.98 & \cellcolor{gray!15}0.40 & 0.34 & 0.70 & 1.33 & \cellcolor{gray!15}0.79 \\

\rowcolor{gray!15} $\operatorname{MomAD\ (Ours)}$ & 
\textbf{0.33}\textit{\fontsize{6}{0}\selectfont\textcolor{red}{\textbf{-0.02}}} & 
\textbf{0.70}\textit{\fontsize{6}{0}\selectfont\textcolor{red}{\textbf{-0.07}}} & 
\textbf{1.24}\textit{\fontsize{6}{0}\selectfont\textcolor{red}{\textbf{-0.22}}} & 
\textbf{0.76}\textit{\fontsize{6}{0}\selectfont\textcolor{red}{\textbf{-0.10}}} & 
0.03\textit{\fontsize{6}{0}\selectfont\textcolor{red}{\textbf{-0.01}}} & 
\textbf{0.13}\textit{\fontsize{6}{0}\selectfont\textcolor{red}{\textbf{-0.04}}} & 
\textbf{0.79}\textit{\fontsize{6}{0}\selectfont\textcolor{red}{\textbf{-0.19}}} & 
\textbf{0.32}\textit{\fontsize{6}{0}\selectfont\textcolor{red}{\textbf{-0.08}}} & 
0.32\textit{\fontsize{6}{0}\selectfont\textcolor{red}{\textbf{-0.02}}} & 
\textbf{0.54}\textit{\fontsize{6}{0}\selectfont\textcolor{red}{\textbf{-0.16}}} & 
\textbf{1.05}\textit{\fontsize{6}{0}\selectfont\textcolor{red}{\textbf{-0.28}}} & 
\textbf{0.63}\textit{\fontsize{6}{0}\selectfont\textcolor{red}{\textbf{-0.16}}} \\
\bottomrule
\end{tabular}}
\label{tab_nuscenes_planning_turning}
\end{table*}

\begin{table}[htp]
\Large
\centering
  \caption{Long trajectory planning results on the $\operatorname{nuScenes}$ and $\operatorname{Turning\text{-}nuScenes}$ validation sets. We train models for 10 epochs for 6s-horizon prediction. $\operatorname{T\text{-}nuScenes}$ indicates the challenging $\operatorname{Turning\text{-}nuScenes}$. We follow the VAD \cite{jiang2023vad} evaluation metric.}\vspace{-1ex}
  \renewcommand\arraystretch{1.0}
  \setlength{\tabcolsep}{1.0mm}
  \resizebox{\linewidth}{!}{
  \begin{tabular}{ll ccc ccc ccc}
\toprule
\multirow{2}{*}{$\operatorname{Split}$} & \multirow{2}{*}{$\operatorname{Method}$} & \multicolumn{3}{c}{$\operatorname{L2\ (m)}\downarrow$} & \multicolumn{3}{c}{$\operatorname{Col.\ Rate\ (\%)}\downarrow$} & \multicolumn{3}{c}{$\operatorname{TPC\ (m)}\downarrow$} \\
\cmidrule(lr){3-5} \cmidrule(lr){6-8} \cmidrule(lr){9-11}
& & 4s & 5s & 6s & 4s & 5s & 6s & 4s & 5s & 6s \\
\midrule
\multirow{3}{*}{$\operatorname{nuScenes}$} 
& $\operatorname{SparseDrive}$~\cite{sun2024sparsedrive} & 1.75 & 2.32 & 2.95 & 0.87 & 1.54 & 2.33 & 1.33 & 1.66 & 1.99 \\
& \cellcolor{gray!15} $\operatorname{MomAD}$ & 
\cellcolor{gray!15}1.67 & \cellcolor{gray!15}1.98 & \cellcolor{gray!15}2.45 & 
\cellcolor{gray!15}0.83 & \cellcolor{gray!15}1.43 & \cellcolor{gray!15}2.13 & 
\cellcolor{gray!15}1.19 & \cellcolor{gray!15}1.45 & \cellcolor{gray!15}1.61 \\
& & 
\textit{\textcolor{blue}{\textbf{-0.09}}} & 
\textit{\textcolor{blue}{\textbf{-0.34}}} & 
\textit{\textcolor{blue}{\textbf{-0.50}}} & 
\textit{\textcolor{blue}{\textbf{-0.04}}} & 
\textit{\textcolor{blue}{\textbf{-0.11}}} & 
\textit{\textcolor{blue}{\textbf{-0.20}}} & 
\textit{\textcolor{blue}{\textbf{-0.14}}} & 
\textit{\textcolor{blue}{\textbf{-0.21}}} & 
\textit{\textcolor{blue}{\textbf{-0.38}}} \\
\midrule
\multirow{3}{*}{$\operatorname{T\text{-}nuScenes}$} 
& $\operatorname{SparseDrive}$~\cite{sun2024sparsedrive} & 2.07 & 2.71 & 3.36 & 0.91 & 1.71 & 2.57 & 1.54 & 2.31 & 2.90 \\
& \cellcolor{gray!15} $\operatorname{MomAD}$ & 
\cellcolor{gray!15}1.80 & \cellcolor{gray!15}2.07 & \cellcolor{gray!15}2.51 & 
\cellcolor{gray!15}0.85 & \cellcolor{gray!15}1.57 & \cellcolor{gray!15}2.31 & 
\cellcolor{gray!15}1.37 & \cellcolor{gray!15}1.58 & \cellcolor{gray!15}1.93 
\\
& & 
\textit{\textcolor{red}{\textbf{-0.27}}} & 
\textit{\textcolor{red}{\textbf{-0.64}}} & 
\textit{\selectfont\textcolor{red}{\textbf{-0.85}}} & 
\textit{\selectfont\textcolor{red}{\textbf{-0.06}}} & 
\textit{\selectfont\textcolor{red}{\textbf{-0.14}}} & 
\textit{\textcolor{red}{\textbf{-0.26}}} & 
\textit{\textcolor{red}{\textbf{-0.17}}} & 
\textit{\textcolor{red}{\textbf{-0.73}}} & 
\textit{\textcolor{red}{\textbf{-0.97}}} \\
\bottomrule
\end{tabular} }
\label{tab_nuscenes_Turning_nuscenes_planning_6s}
\end{table}

\begin{table}[htp]
% \tiny
\centering
\addtolength{\tabcolsep}{0.1pt}
\caption{$\operatorname{Open-loop}$ and $\operatorname{Closed-loop}$ results on $\operatorname{Bench2Drive}$ (V0.0.3) under base training set.
% Avg. L2 is averaged over the predictions in 2 seconds under 2Hz, similar to UniAD. 
`mmt' refers multi-modal trajectory variant of $\operatorname{VAD}$ and $^*$ the re-implementation. 
% `DS' denotes Driving Score. `SR' denotes Success Rate. `Effi' denotes Efficiency. `Comf' denotes Comfortness.
}
     \vspace{-1em}
  \renewcommand\arraystretch{0.9}
  \tabcolsep=0.3mm %%%%%%%%%
  \resizebox{\linewidth}{!}{
  \begin{tabular}{lccccc}
    \toprule
   \multirow{2}{*}{\textbf{$\operatorname{Method}$}}& \textbf{$\operatorname{Open-loop\ Metric}$}     & \multicolumn{4}{c}{\textbf{$\operatorname{Closed-loop\ Metric}$}}  \\
   \cmidrule(lr){2-2} \cmidrule(lr){3-6} 
   &$\operatorname{Avg.\ L2}\downarrow$&$\operatorname{DS}\uparrow$ & $\operatorname{SR\ (\%)}\uparrow$ & $\operatorname{Effi}\uparrow$  & $\operatorname{Comf}\uparrow$ \\
\midrule

$\operatorname{VAD}$ & 0.91& 42.35& 15.00& 157.94& 46.01 \\
${\operatorname{VAD}_{\operatorname{mmt}}}^{*}$ & 0.89& 42.87& 15.91& 158.12& 47.22 \\
$\operatorname{MomAD\ (Euclidean)}$ & 0.87& 44.22& 16.91& 161.77& 48.70 \\
\cellcolor{gray!15}$\operatorname{MomAD}$ & \cellcolor{gray!15}\textbf{0.85}& \cellcolor{gray!15}\textbf{45.35}&\cellcolor{gray!15} \textbf{17.44}& \cellcolor{gray!15}\textbf{162.09}& \cellcolor{gray!15}\textbf{49.34} \\
\midrule

${\operatorname{SparseDrive}}^{*}$ & 0.87& 44.54& 16.71& 170.21& 48.63 \\
$\operatorname{MomAD\ (Euclidean)}$ & 0.84& 46.12& 17.45& 173.35& 50.98 \\

\cellcolor{gray!15}$\operatorname{MomAD}$ & \cellcolor{gray!15}\textbf{0.82}& \cellcolor{gray!15}\textbf{47.91}& \cellcolor{gray!15}\textbf{18.11}& \cellcolor{gray!15}\textbf{174.91}& \cellcolor{gray!15}\textbf{51.20} \\
\bottomrule
\end{tabular} }
\label{tab_b2d}
\end{table}

\begin{table*}[htp]
\Large
\centering
  \caption{Perception and motion results on the \textbf{nuScenes} validation dataset. $^{\dagger}$ indicates the results are reproduced with the official checkpoint. $\operatorname{AP}_{d}$ denotes $\operatorname{AP}_{\operatorname{divider}}$. $\operatorname{AP}_{b}$ denotes $\operatorname{AP}_{\operatorname{boundary}}$. $\operatorname{mADE}$ denotes $\operatorname{minADE}$. $\operatorname{mFDE}$ denotes $\operatorname{minFDE}$.}
  \renewcommand\arraystretch{1.0}
  \setlength{\tabcolsep}{0.5mm} % Adjust column spacing
  \resizebox{\linewidth}{!}{
  \begin{tabular}{lccccccc cccc cccc cccc}
\toprule
\multirow{2}{*}{$\operatorname{Method}$} & \multicolumn{7}{c}{$\operatorname{3D\ Object\ Detection}$} & \multicolumn{4}{c}{$\operatorname{Multi-Object\ Tracking}$} & \multicolumn{4}{c}{$\operatorname{Online\ Mapping}$} & \multicolumn{4}{c}{$\operatorname{Motion\ Prediction}$} \\
\cmidrule(lr){2-8} \cmidrule(lr){9-12} \cmidrule(lr){13-16} \cmidrule(lr){17-20}
& $\operatorname{mAP}\uparrow$ & $\operatorname{NDS}\uparrow$ & $\operatorname{mATE}\downarrow$ & $\operatorname{mASE}\downarrow$ & $\operatorname{mAOE}\downarrow$ & $\operatorname{mAVE}\downarrow$ & $\operatorname{mAAE}\downarrow$ 
& $\operatorname{AMOTA}\uparrow$ & $\operatorname{AMOTP}\downarrow$ & $\operatorname{Recall}\uparrow$ & $\operatorname{IDS}\downarrow$
& $\operatorname{mAP}\uparrow$ & $\operatorname{AP}_{\operatorname{ped}}\uparrow$ & $\operatorname{AP}_{d}\uparrow$ & $\operatorname{AP}_{b}\uparrow$ 
& $\operatorname{mADE}\downarrow$ & $\operatorname{mFDE}\downarrow$ & $\operatorname{MR}\downarrow$ & $\operatorname{EPA}\uparrow$ \\
\midrule
$\operatorname{UniAD}$~\cite{uniad} & \cellcolor{gray!15}0.380 & \cellcolor{gray!15}0.498 & 0.684 & 0.277 & 0.383 & 0.381 & 0.192 
& \cellcolor{gray!15}0.359 & 1.320 & 0.467 & 906 
& \cellcolor{gray!15}- & - & - & - 
& \cellcolor{gray!15}0.71 & 1.02 & 0.151 & 0.456 \\

$\operatorname{VAD}^{\dagger}$~\cite{jiang2023vad} & \cellcolor{gray!15}0.312 & \cellcolor{gray!15}0.435 & 0.610 & 0.288 & 0.541 & 0.534 & 0.228 
& \cellcolor{gray!15}- & - & - & - 
& \cellcolor{gray!15}47.6 & 40.6 & 51.5 & 50.6 
& \cellcolor{gray!15}- & - & - & - \\

$\operatorname{SparseDrive}$~\cite{sun2024sparsedrive} & \cellcolor{gray!15}0.418 & \cellcolor{gray!15}0.525 & 0.566 & 0.275 & 0.552 & 0.261 & 0.190 
& \cellcolor{gray!15}0.386 & 1.254 & 0.499 & 886 
& \cellcolor{gray!15}55.1 & 49.9 & 57.0 & 58.4 
& \cellcolor{gray!15}0.62 & 0.99 & \textbf{0.136} & 0.482 \\

\cellcolor{gray!15}$\operatorname{MomAD\ (Ours)}$ & \cellcolor{gray!15}\textbf{0.423} & \cellcolor{gray!15}\textbf{0.531} & \cellcolor{gray!15}\textbf{0.561} & \cellcolor{gray!15}\textbf{0.269} & \cellcolor{gray!15}\textbf{0.549} & \cellcolor{gray!15}\textbf{0.258} & \cellcolor{gray!15}\textbf{0.188} 
& \cellcolor{gray!15}\textbf{0.391} & \cellcolor{gray!15}\textbf{1.243} & \cellcolor{gray!15}\textbf{0.509} & \cellcolor{gray!15}\textbf{853} 
& \cellcolor{gray!15}\textbf{55.9} & \cellcolor{gray!15}\textbf{50.7} & \cellcolor{gray!15}\textbf{58.1} & \cellcolor{gray!15}\textbf{58.9} 
& \cellcolor{gray!15}\textbf{0.61} & \cellcolor{gray!15}\textbf{0.98} & \cellcolor{gray!15}0.137 & \cellcolor{gray!15}\textbf{0.499} \\

\bottomrule
\end{tabular} }
\label{tab_nuscenes_perception_motion}
\end{table*}

\section{Experiments} 
\label{experiments}
% \subsection{Experiment Setup}
\subsection{Experimental Setup}

\textbf{Datasets.} We conducted extensive experiments on the widely adopted \textbf{nuScenes} dataset~\cite{nuscenes} to evaluate tasks including detection, online mapping and planning in an open-loop setting. The nuScenes dataset comprises 1,000 driving scenes, with 700 and 150 sequences allocated for training and validation. Each scene spans around 20 seconds and contains roughly 40 key-frames annotated at 2Hz, where each sample includes six images captured by surrounding cameras covering 360° FOV horizontally and point clouds collected by both LiDAR and radar sensors. Since most planning tasks in nuScenes focus predominantly on go-straight commands, we curate a challenging subset of turning scenarios to form the \textbf{Turning-nuScenes} dataset, aimed at verifying the temporal consistency of predicted trajectories across time steps. Planning samples for turns are selected by setting the threshold between 3s and 0.5s of `gt\_ego\_fut\_trajs' to 25. The turning nuScenes validation dataset constitutes only one-tenth of the full nuScenes validation set, including 17 scenes with 680 samples. We use \textbf{Bench2Drive} \cite{bench2drive}, a closed-loop evaluation protocol under CARLA Leaderboard 2.0 for end-to-end autonomous driving. It provides an official training set, where we use the base set (1000 clips) for fair comparison with all the other baselines. We use the official 220
routes for evaluation.

\noindent\textbf{Evaluation Metrics for Planning.} For planning evaluation, we adopt the commonly used L2 Displacement Error (L2) and Collition Rate to assess planning performance. The calculation of the L2 error follows VAD~\cite{jiang2023vad} and the collision rate is aligned with SparseDrive~\cite{sun2024sparsedrive}. However, the mainstream planning metrics cannot faithfully reveal the stability of predicted trajectories. Therefore, we introduce a novel metric, Trajectory Prediction Consistency  ($\operatorname{TPC}$), to measure the disparity between current predicted trajectories and historical predicted trajectories, allowing for a more comprehensive assessment of the consistency of trajectories. With coordinates transformed, the $\operatorname{TPC}$ between the current predicted trajectory $T_{t}^{\operatorname{Pred}}$ and historical one $T_{t-1}^{\operatorname{Pred}}$ is defined as,
\begin{align}
    \operatorname{TPC} = \frac{1}{N_{T}} \sqrt{ \sum_{i=1}^{N} \left( \left( T_{t}^{\operatorname{Pred}} - T_{t-1}^{\operatorname{Pred}}  \right)^2 \cdot T^{\operatorname{Mask}}_{\operatorname{GT}} \right)}, 
\end{align}
where $N_{T}$ is the total number of GT trajectories in the validation set, and $T^{\operatorname{Mask}}_{\operatorname{GT}}$ is the mask for trajectories exceeding the overlapped time period of two trajectories.
Our $\operatorname{TPC}$ metric evaluates whether autonomous vehicles adhere to predicted trajectories, ensuring continuity across frames. Notably, the $\operatorname{TPC}$ metric provides a statistical perspective on the dataset-wide evaluation rather than at the individual sample level.

\begin{table}[htp]
\Large
\centering
  \caption{
  Ablation studies of the sparse perception module in MomAD on the nuScenes validation split. 
  The Encoder and Denoise ($\sigma$) module is denoted as $\operatorname{ED}$.  $\operatorname{MP}$ represents Momentum planning. $\operatorname{NS}$ is the Gaussian noise factor controlling the noise level. Noise is applied during training only.  We follow the VAD \cite{jiang2023vad} evaluation metric.}
  \renewcommand\arraystretch{1.0}
  \setlength{\tabcolsep}{.5mm} % Adjust column spacing
  \resizebox{\linewidth}{!}{
  \begin{tabular}{lllcccccccccc}
\toprule
\multirow{2}{*}{$\operatorname{ED}$} & \multirow{2}{*}{$\operatorname{MP}$} & \multirow{2}{*}{$\operatorname{NS}$} & \multicolumn{2}{c}{$\operatorname{Detection}$} & $\operatorname{Tracking}$ & \multicolumn{2}{c}{$\operatorname{Online\, Mapping}$} & $\operatorname{Motion}$ & \multicolumn{3}{c}{$\operatorname{Planning\ (Avg.)}$} \\
\cmidrule(lr){4-5} \cmidrule(lr){6-6} \cmidrule(lr){7-8} \cmidrule(lr){9-9} \cmidrule(lr){10-12}
&&& $\operatorname{mAP} \uparrow$ & $\operatorname{NDS} \uparrow$ & $\operatorname{AMOTA} \uparrow$ & $\operatorname{mAP} \uparrow$ & $\operatorname{AP}_{\operatorname{ped}} \uparrow$ & $\operatorname{mADE} \downarrow$ & $\operatorname{L2} \downarrow$ & $\operatorname{Col.} \downarrow$ & $\operatorname{TPC} \downarrow$ \\
\midrule
& & 0.0          & 0.407 & 0.521 & 0.381 & 55.0 & 49.3 & 0.63 & 0.62 & 0.14 & 0.56 \\ 
& \Checkmark & 0.0          & 0.405 & 0.520 & 0.380 & 55.1 & 49.5 & 0.63 & 0.61 & 0.13 & 0.55 \\ 
\Checkmark & & 0.1          & 0.420 & 0.530 & 0.390 & 55.8 & 50.5 & 0.58 & 0.61 & 0.12 & 0.55 \\

\Checkmark & \Checkmark & 0.0          & 0.417 & 0.528 & 0.386 & 55.4 & 50.6 & 0.63 & 0.61 & 0.11 & 0.55 \\

\Checkmark & \Checkmark & 0.05         & 0.421 & 0.529 & 0.388 & 55.6 & \textbf{50.8} & 0.62 & 0.61 & 0.11 & 0.54 \\

\rowcolor{gray!15} \Checkmark & \Checkmark & 0.1 & \textbf{0.423} & \textbf{0.531} & \textbf{0.391} & \textbf{55.9} & 50.7 & \textbf{0.61} & \textbf{0.60} & \textbf{0.09} & \textbf{0.54} \\

\Checkmark & \Checkmark & 0.2          & 0.418 & 0.520 & 0.388 & 54.4 & 49.2 & 0.63 & 0.62 & 0.18 & 0.58 \\

\Checkmark & \Checkmark & 0.3          & 0.412 & 0.518 & 0.383 & 54.0 & 48.8 & 0.65 & 0.64 & 0.22 & 0.61 \\

\bottomrule
\end{tabular} }
\label{tab_nuscenes_ablation_sparse_percetion}
\end{table}

\begin{table}[htp]
\normalsize
\centering
  \caption{Impact of history frames in MomAD on the Turning-nuScenes validation set. %Momentum planning is denoted as `\textbf{MP}'. 
  $t$ denotes the frame number, where $1$ indicates the history is empty (represented by $0$), $2$ signifies that the historical result corresponds to the previous 1 frame, and $3$ indicates that the historical result pertains to the previous 2 frames.  We follow the VAD \cite{jiang2023vad} evaluation metric.}
  \renewcommand\arraystretch{1.0}
  \setlength{\tabcolsep}{1.45mm} % Adjust column spacing
  \resizebox{\linewidth}{!}{
  \begin{tabular}{llllccc ccc ccc}
\toprule
\multirow{2}{*}{$\operatorname{ED}$} & \multirow{2}{*}{$\operatorname{MP}$} & \multirow{2}{*}{$\operatorname{NS}$} & \multirow{2}{*}{$t$} & \multicolumn{3}{c}{$\operatorname{L2 (m)} \downarrow$} & \multicolumn{3}{c}{$\operatorname{Col.\ Rate (\%)} \downarrow$} & \multicolumn{3}{c}{$\operatorname{TPC (m)} \downarrow$} \\
\cmidrule(lr){5-7} \cmidrule(lr){8-10} \cmidrule(lr){11-13}
&&& & 2s & 3s & Avg. & 2s & 3s & Avg. & 2s & 3s & Avg. \\
\midrule
&&& & 0.77 & 1.46 & 0.86 & 0.17 & 0.98 & 0.40 & 0.70 & 1.33 & 0.79 \\
&\Checkmark & & 2 & 0.71 & 1.27 & 0.77 & 0.14 & 0.83 & 0.34 & 0.55 & 1.07 & 0.65 \\
\Checkmark & & 0.1 & & 0.77 & 1.45 & 0.86 & 0.18 & 0.97 & 0.39 & 0.69 & 1.33 & 0.79 \\ 
\Checkmark & \Checkmark & 0.1 & 1 & 0.78 & 1.48 & 0.88 & 0.18 & 0.98 & 0.39 & 0.70 & 1.35 & 0.81 \\ 
\rowcolor{gray!15} \Checkmark & \Checkmark & 0.1 & 2 & \textbf{0.70} & \textbf{1.24} & \textbf{0.76} & \textbf{0.13} & \textbf{0.79} & \textbf{0.32} & \textbf{0.54} & \textbf{1.05} & \textbf{0.63} \\
\Checkmark & \Checkmark & 0.1 & 3 & 0.72 & 1.27 & 0.78 & 0.14 & 0.84 & 0.35 & 0.56 & 1.09 & 0.66 \\
\bottomrule
\end{tabular} }
\label{tab_nuscenes_ablation_planning_turning}
\end{table}

\begin{figure*}[t]
    \centering
    \includegraphics[width=1.0\linewidth]{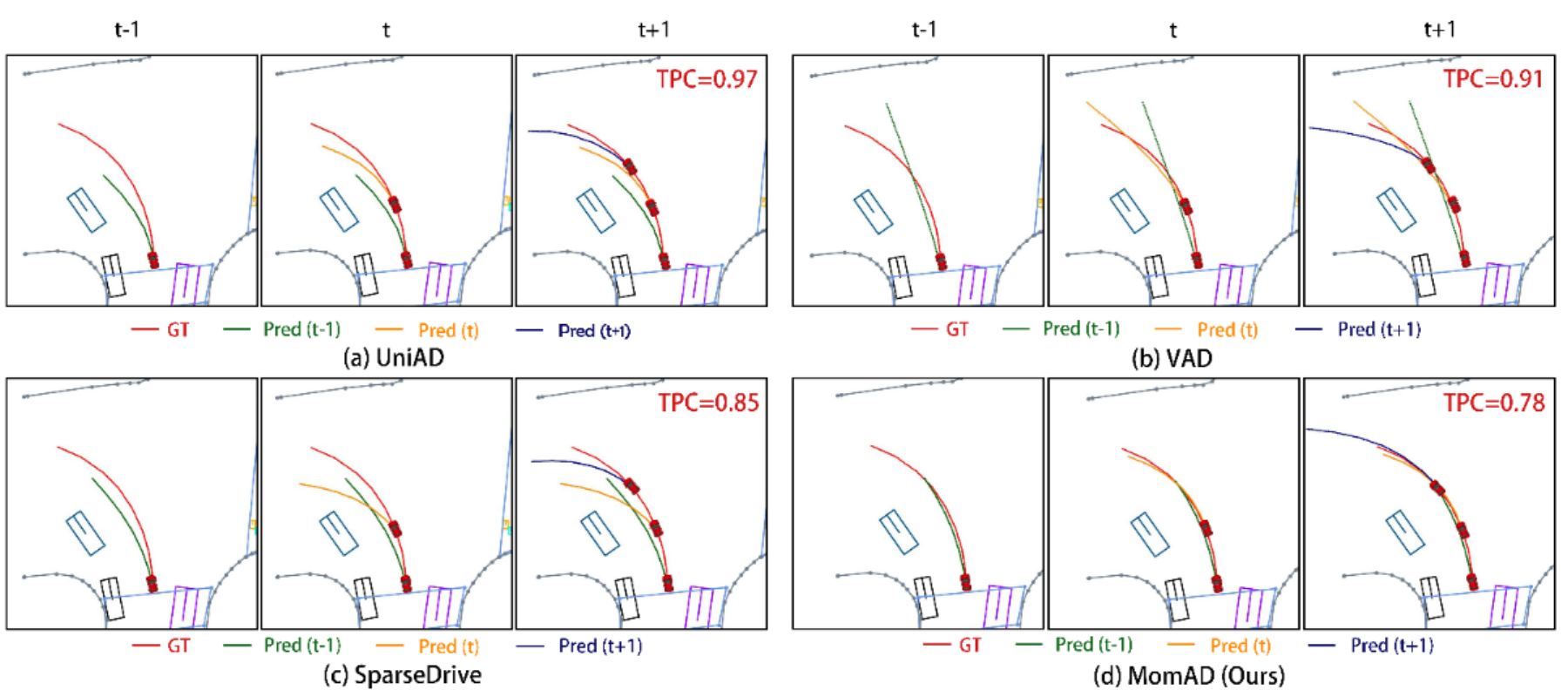}
    \caption{Visualization results of MomAD compared with UniAD, VAD and SparseDrive across multiple frames. The proposed MomAD achieves temporal consistency whichever from the predicted trajectory compared with ground truth (GT) or from the TPC metric.}
    \label{fig:vis}
\end{figure*}

\subsection{Main Results}
% We compared our MomAD with SOTA end-to-end autonomous driving methods showed in Table \ref{tab_nuscenes_planning}. 
% Among end-to-end methods, our MomAD has surpassed previous SOTAs in end-to-end tasks incuding 3D object detection, multi-object tracking, online mapping, motion prediction, and planning.
To conduct a comprehensive comparison, we have undertaken an exhaustive analysis of two distinct metrics presented in Table \ref{tab_nuscenes_planning}, which are derived from UniAD~\cite{uniad} and VAD~\cite{jiang2023vad}. It is noteworthy that, aside from Table \ref{tab_nuscenes_planning}, all other tables rely on the VAD~\cite{jiang2023vad} evaluation metrics for their assessments.
\subsubsection{Planning Results}

\textbf{NuScenes.} As shown in Table~\ref{tab_nuscenes_planning}, MomAD achieves an L2 error of 0.60m, a collision rate of 0.09\%, and a TPC of 0.54m, respectively. %Our method achieves SOTA performance in the TPC metric, which demonstrates that we can solve the temporal inconsistency issue and ensure more stable driving for an ego vehicle.
Compared to SOTAs like UniAD~\cite{uniad}, VAD~\cite{jiang2023vad} and SparseDrive \cite{sun2024sparsedrive}, our method shows SOTA performance in planning results. It is worth noting that we have made significant improvements in TPC, which directly proves our effectiveness in timing consistency. It is worth noting that we achieved significant improvements in TPC at 0.30m, 0.53m, and 0.78m at 1s, 2s, and 3s on the nuScenes dataset, directly demonstrating our effectiveness in temporal consistency. Additionally, our MomAD is straightforward and achieves an FPS of 7.8, %second only to SparseDrive.
a slightly slower than SparseDrive.
In summary, our MomAD effectively utilizes the smoothing advantage of Momentum and has a significant effect on improving temporal consistency.

\noindent\textbf{Turning-nuScenes.} 
% As pointed out in Ref. ~\cite{bevplanner}, the nuScenes dataset includes many straight routes, which cannot fully evaluate the true value of end-to-end methods. However, the simplicity of straight-line paths may obscure the model's actual performance in complex scenarios. Therefore, we evaluated our MomAD in on the Turing-nuScenes validation set, as shown in Table ~\ref{tab_nuscenes_planning_turning}, where it demonstrates significant advantages. 
% SparseDrive ~\cite{sun2024sparsedrive} is a SOTA end-to-end method utilizing multi-modal trajectories. Although it performs well in various scenarios, it struggles in situations that require maintaining driving stability, such as during turns. However, our method demonstrates superior performance in trajectory prediction consistency, as indicated by the TPC metric. This means that MomAD not only provides effective trajectory predictions under standard conditions but also maintains high consistency and reliability when facing dynamic changes and complex environments.
As noted in \cite{bevplanner}, the nuScenes dataset features many straight routes, which limits the assessment of end-to-end methods. The simplicity of these paths can mask a model's true performance in complex scenarios. To address this, we evaluated our MomAD on the Turing-nuScenes validation set, as shown in Table~\ref{tab_nuscenes_planning_turning}. % where the results demonstrate the significant advantages of MomAD. 
SparseDrive~\cite{sun2024sparsedrive}, a SOTA end-to-end method utilizing multi-modal trajectories, performs well across various scenarios but struggles with driving stability during turns. In contrast, MomAD exhibits superior consistency in trajectory predictions, as indicated by the TPC metric. Overall, MomAD not only delivers effective trajectory predictions under standard conditions but also maintains reliability amid dynamic changes and complex environments.

\noindent\textbf{Long Trajectory Prediction.} Accurate long trajectory prediction is vital for enhancing the stability of autonomous driving, is useful to evaluate molels' ability to address temporal inconsistency issues in multi-modal trajectory planning. As shown in Table~\ref{tab_nuscenes_Turning_nuscenes_planning_6s}, we compared SparseDrive and MomAD in 4-6s long trajectory prediction on the nuScenes and Turning-nuScenes dataset, demonstrating a significant performance improvement. Specifically, in nuScenes dataset,  compared with SparseDrive, MomAD experienced a decrease of 0.09m (5.14\%), 0.34m (14.66\%), and 0.50m (16.95\%) in the 4s, 5s, and 6s of L2 error, a decrease of 0.04\%, 0.11\%, and 0.20\% in the 4s, 5s, and 6s of collision rate, and a decrease of 0.14m (10.53\%), 0.21m (12.65\%), and 0.38m (19.10\%) in the 4s, 5s, and 6s of TPC, respectively. Futhermore, in Turning-nuScenes dataset,  compared with SparseDrive, MomAD experienced a decrease of 0.27m (13.04\%), 0.64m (23.62\%), and 0.85m (25.30\%) in the 4s, 5s, and 6s of L2 error, a decrease of 0.06\%, 0.14\%, and 0.26\% in the 4s, 5s, and 6s of collision rate, and a decrease of 0.17m (11.04\%), 0.73m (31.60\%), and 0.97m (32.45\%) in the 4s, 5s, and 6s of TPC, respectively. We can observe that MomAD significantly improves trajectory predictions at farther distances, with a magnitude improvement at 6s. In summary, our MomAD has improved the performance of long trajectory predictions, which further proves that MomAD can effectively alleviate the problem of temporal inconsistency.

\noindent\textbf{Bench2Drive.} 
We have included evaluations on the challenging closed-loop results on Bench2Drive dataset, as shown in Table \ref{tab_b2d}, w which covers 44 interactive scenes (\textit{e.g.}, cut-ins, overtaking, detours) and 220 routes across diverse weather conditions and locations. Our MomAD improves success rate by \textbf{16.3\%} and \textbf{8.4\%} over the VAD multi-modal  variant and SparseDrive, and enhances the Comfortness score (trajectory smoothness) by \textbf{7.2\%} and \textbf{5.3\%}, demonstrating its effectiveness.

\subsubsection{Perception and Motion Prediction Results}
Sparse representation is efficient but suffers from instability issues caused by the variability of instance features. To address these issues, we have enhanced the instance features using the Encoder and Denoise ($\sigma$) module (denoted as \textbf{ED}) within the sparse perception framework, ensuring end-to-end stability for autonomous driving. As shown in Table~\ref{tab_nuscenes_perception_motion}, our MomAD perception module includes 3D object detection, multi-object tracking, and online mapping tasks. For 3D object detection, MomAD achieves 42.3\% mAP and 53.1\% NPS, improving the mAP by 0.5\% and the NDS by 0.6\% compared to the baseline SparseDrive~\cite{sun2024sparsedrive}. For multi-object tracking, MomAD achieves an AMOTA of 39.1\%, surpassing the baseline SparseDrive by 0.5\%. For online mapping, compared to 55.1\% mAP for the baseline SparseDrive, our MomAD achieves 55.9\% mAP, improving the mAP by 0.8\%.
For motion prediction, our MomAD outperforms SparseDrive ~\cite{sun2024sparsedrive} and UniAD ~\cite{uniad}, %. Specifically, our MomAD 
achieves better motion prediction performance by considering the influence of an ego vehicle on other agents. In detail, Our MomAD achieves a 0.61m minADE, 0.98 minFDE, and 13.7\% MR, and 0.499 EPA, respectively. %Overall, MomAD has achieved better performance in motion prediction task.

\begin{table}[t]
\scriptsize
\centering
  \caption{Ablation studies of the impact of the different modules
  in $\operatorname{MP}$ on the Turning-nuScenes validation dataset. $\operatorname{QM}$ denotes Long-horizon Query Mixer, $\operatorname{TP}$ denotes Trajectory Predictor. $\operatorname{Add}$ refers to the addition operation between the historical planning query $\mathbf{Q}_{t-1}^p$ and the selected planning query $\mathbf{Q}_{t}^{p*}$.  We follow the VAD \cite{jiang2023vad} evaluation metric.}
  \renewcommand\arraystretch{1.0}
  \setlength{\tabcolsep}{1.2mm} % Adjust column spacing
  \resizebox{\linewidth}{!}{
  \begin{tabular}{cccccc ccc ccc}
\toprule
\multirow{2}{*}{$\operatorname{QM}$} & \multirow{2}{*}{$\operatorname{Add}$} & \multirow{2}{*}{$\operatorname{TP}$} & \multicolumn{3}{c}{$\operatorname{L2\ (m)} \downarrow$} & \multicolumn{3}{c}{$\operatorname{Col.\ Rate\ (\%)} \downarrow$} & \multicolumn{3}{c}{$\operatorname{TPC\ (m)} \downarrow$} \\
\cmidrule(lr){4-6} \cmidrule(lr){7-9} \cmidrule(lr){10-12}
&&& 2s & 3s & Avg. & 2s & 3s & Avg. & 2s & 3s & Avg. \\
\midrule
& & & 0.77 & 1.46 & 0.86 & 0.17 & 0.98 & 0.40 & 0.70 & 1.33 & 0.79 \\

& \Checkmark & \Checkmark & 0.76 & 1.38 & 0.82 & 0.14 & 0.88 & 0.36 & 0.62 & 1.21 & 0.67 \\

\rowcolor{gray!15} \Checkmark & & \Checkmark & \textbf{0.70} & \textbf{1.24} & \textbf{0.76} & \textbf{0.13} & \textbf{0.79} & \textbf{0.32} & \textbf{0.54} & \textbf{1.05} & \textbf{0.63} \\
\bottomrule
\end{tabular} }
\label{tab_nuscenes_ablation_MP_planning_turning}
\end{table}

\subsection{Ablation Study}
% We conduct extensive experiments to study the effectiveness and necessity of each module in our MomAD.

\noindent\textbf{Roles of `ED' module in Sparse Perception.}
 Sparse representation end-to-end methods yield efficient computation but unstable metrics. Further details are available in the Appendix. As shown in Table~\ref{tab_nuscenes_ablation_sparse_percetion}, the Encoder and Denoise ($\sigma$) module (\textbf{ED}) within sparse perception enhances the instance features,  significantly impacts the overall pipeline. By introducing Gaussian noise and employing techniques, % to mitigate it, 
 the robustness of instance features is improved, particularly when training with Noisy at 0.1. Our findings suggest that a controlled level of noise can enhance the end-to-end capabilities of sparse methods during training, offering insights for the community on sparse end-to-end methods.

\noindent\textbf{Roles of `MP' module in Planing.}
As Li et al.~\cite{bevplanner} have stated, most end-to-end autonomous driving methods perform poorly in turning scenarios.
As shown in Table~\ref{tab_nuscenes_ablation_planning_turning}, to better evaluate the planning performance of end-to-end methods in turning scenarios, our MomAD is evaluated on the Turning-nuScenes validation dataset rather than only on the full nuScenes validation dataset. Specifically, under the premise of executing `ED', at $t=1$, providing a 0 value to the MP module does not improve performance. We have tried to change the MLP operation of the planning to a more complex operation, but it does not enhance the results. However, when $t=2$, historical queries and results are used, L2 (Avg) reaches 0.76m, Col. (Avg) reaches 0.32 \%, and TPC reaches 0.63m, which represents a significant improvement. In addition, when $t=3$, more frames are fused, the improvement has actually decreased, which may be due to the uncertainty introduced by the departure of historical features, but there is still an overall improvement.

\noindent\textbf{Roles of different sub-modules in `MP' module.}
As shown in Table \ref{tab_nuscenes_ablation_MP_planning_turning}, we conducted an in-depth analysis of the internal mechanism of momentum planning. We found that simply using the native `Add' operation to regenerate the planning results can achieve a good improvement, with L2 (Avg.), Col. (Avg.), and TPC (Avg.) decreasing by 0.04m, 0.04\%, and 0.12m, respectively. However, the `Add' operation alone does not fully utilize historical features. Our Long-horizon Query Mixer has achieved the current optimal performance. Overall, historical results are very important for current outcomes, and their reasonable utilization can maximize the performance of end-to-end planning.

\subsection{Visualization} 
% \noindent\textbf{Qualitative Study of Planning Results.} 
As shown in Figure \ref{fig:vis}, we showcase multi-frame qualitative comparisons of end-to-end solutions, including UniAD \cite{uniad}, VAD \cite{jiang2023vad}, SparseDrive \cite{sun2024sparsedrive}, and the proposed MomAD. In a representative turning scenario, the MomAD approach demonstrates superior long-term awareness of surrounding vehicles, reducing the likelihood of collisions. Additionally, it generates smoother ego-vehicle trajectories (shown in yellow and blue) that closely align with the ground-truth trajectory (in red). This highlights its strong temporal consistency and lower TPC scores. 
Additional visualizations for various driving commands are provided in the Appendix.

\section{Conclusion and Future Work} 
The proposed MomAD framework addresses key challenges in planning stability and robustness for end-to-end autonomous driving systems. By leveraging trajectory momentum and perception momentum, MomAD stabilizes trajectory predictions through Topological Trajectory Matching (TTM) and Momentum Planning Interactor (MPI), ensuring temporal coherence and enriching long-horizon context. Evaluations on nuScenes and the curated Turning-nuScenes validation set demonstrate its superior performance in reducing collision rates and improving trajectory consistency compared to state-of-the-art methods. While MomAD improves temporal consistency in long-horizon trajectory prediction, a gap remains due to mode collapse induced by the standard teacher-forcing approach to trajectory regression, limiting trajectory diversity. Future work will explore diffusion models and speculative decoding to enhance trajectory diversity while balancing efficiency. 

\section*{Acknowledgements}
We sincerely appreciate the helpful discussions provided by Wenchao Sun, Bo Jiang, Bencheng Liao from Horizon Robotics. This work was supported by the National Key R\&D Program of China (2018AAA0100302).

\small
\bibliographystyle{ieeenat_fullname}
\bibliography{main}

\newpage
\clearpage
\maketitlesupplementary
\appendix
\setcounter{page}{1}
\setcounter{section}{0}
\setcounter{figure}{0}
\setcounter{table}{0}
\setcounter{equation}{0}
\renewcommand{\thefigure}{A\arabic{figure}}

\section{Appendix}
\noindent This supplementary material provides additional descriptions of the proposed MomAD framework, including the following supplementary material:
\begin{itemize}
    \item \textbf{\cref{sec:Contribution}:} Summary of contributions.
    \item \textbf{\cref{sec:Turning_dataset}:} The details of Turning-nuScenes dataset.
    \item \textbf{\cref{sec:Implementation_Details}:}  Implementation details.
    \item \textbf{\cref{sec:Results}:} More planning results.
    \item \textbf{\cref{sec:Robustness}:} Detailed Result Analysis on Robustness.
    \item \textbf{\cref{sec:Qualitative}:} More visualizations of planning results.
\end{itemize}

% \subsection{Broader Impact}
% The proposed MomAD framework innovatively enhances end-to-end autonomous driving by introducing trajectory and perception momentum. This approach stabilizes trajectory predictions over time through key components: Topological Trajectory Matching (TTM), which aligns current trajectories with historical paths to maintain consistency, and the Momentum Planning Interactor (MPI), which leverages historical queries to enrich perception and extend context awareness. This dual approach results in improved long-horizon trajectory prediction, reduced collision risks, and better resilience to noise, outperforming existing state-of-the-art methods.
% From a societal perspective, MomAD solves the problem of temporal consistency, leading to safer and more reliable autonomous vehicles. Such improvements can reduce traffic incidents, enhance mobility for those unable to drive, and contribute to more efficient transportation systems, reinforcing the transformative impact of autonomous technology.

\subsection{Contributions}
\label{sec:Contribution}
Our contributions are summarized below.

% \noindent 1) \textbf{Momentum for Planning Concept.} Momentum planning leverages the trajectory and perception momentum to enhance current planning through historical guidance to overcome temporal inconsistency.

\noindent 1) \textbf{MomAD Framework.} We propose MomAD, an end-to-end autonomous driving framework that employs momentum planning. Momentum planning leverages trajectory and perception momentum to enhance current planning through historical guidance, overcoming temporal inconsistency. It addresses key challenges in planning stability and robustness for end-to-end autonomous driving systems.

\noindent 2) \textbf{TTM and MPI.} We propose the Topological Trajectory Matching (TTM) module, which utilizes the Hausdorff Distance to align candidate trajectories with past paths, ensuring temporal coherence and reducing abrupt trajectory changes. Furthermore, we propose the Momentum Planning Interactor (MPI) module. By cross-referencing current and past trajectory data, this module expands the system's perceptual awareness over time, enhancing long-horizon prediction and reducing collision risks.

% \noindent 3) \textbf{Topological Trajectory Matching (TTM).} This component utilizes the Hausdorff Distance to align candidate trajectories with past paths, ensuring temporal coherence and reducing abrupt trajectory changes.

% \noindent 4) \textbf{Momentum Planning Interactor (MPI).} By cross-referencing current and past trajectory data, this module expands the system's perceptual awareness over time, enhancing long-horizon prediction and reducing collision risks.

\noindent 3) \textcolor{blue}{\textbf{New$^*$}} Turning-nuScenes Validation Dataset. We create the Turning-nuScenes val dataset, derived from the nuScenes full validation dataset. This new dataset focuses on turning scenarios, providing a specialized benchmark for evaluating the performance of autonomous driving systems in complex driving situations.

\noindent 4)  \textcolor{blue}{\textbf{New$^*$}} Trajectory Prediction Consistency (TPC) Metric. We introduce the TPC metric to quantitatively assess the consistency of trajectory predictions in existing end-to-end autonomous driving methods, addressing a critical gap in the evaluation of trajectory planning.

% \noindent 5) \textbf{Performance Evaluation.} Through extensive experiments on the nuScenes dataset, we demonstrate that MomAD significantly outperforms SOTA methods in terms of trajectory consistency and stability, highlighting its effectiveness in tackling challenges within autonomous driving planning. We evaluated the results of long trajectory predictions, specifically at 4, 5, and 6 seconds, which are critical for ensuring the stability of autonomous driving systems.

\subsection{The Detail of Turning-nuScenes dataset}
\label{sec:Turning_dataset}
When turning, vehicles need to quickly and accurately adjust their direction, making turning scenarios particularly challenging for the model's ability to maintain stable planning. However, there is currently no dataset specifically designed for evaluating models in turning scenarios. Based on the nuScenes val dataset, we selectively extracted data involving the ego vehicle in turning situations from the validation set to create the Turning-nuScenes dataset. 

\noindent 1) \textbf{Preparation Work.} We extract the data information from the \textit{val} dataset based on the annotations of NuScenes dataset. Specifically, we establish a correspondence between \textit{sample\_token} (the unique identifier of each sample) and \textit{scene\_token} (the unique identifier of each scene) grounded in the provided data annotation information as illustrated in formula \ref{eq:data_prepare}. We also extracted the future trajectory $T_{\operatorname{fut}}$ of the ego vehicle for each sample in the validation dataset over the next three seconds.

\begin{equation}
    \label{eq:data_prepare}
    \begin{aligned}
    dict_{sa}^{sc}[sample\_token] = scene\_token\\
    \end{aligned}
\end{equation}

\noindent 2) \textbf{Sample Select.} Considered that the ego vehicle's driving direction aligns with the y-axis of the world coordinate system, significant changes in the x-coordinate will occur during turns. Thus, we assess potential future turns of the ego vehicle based on changes in its x-coordinate, recording the unique identifier of each sample (\textit{sample\_token}). The specific criteria for judgment are as outlined in the formula \ref{eq:turn_judge},
\begin{equation}
\label{eq:turn_judge}
\begin{cases}
 S_{T} \  \ |T_{\operatorname{fut}}[0] -  T_{\operatorname{fut}}[5]| \ge \varepsilon \\
 S_{S} \  \ |T_{\operatorname{fut}}[0] -  T_{\operatorname{fut}}[5]| < \varepsilon\\
\end{cases}
\end{equation}
where $S_{T}$ and $S_{S}$ represent the states of the ego vehicle during turning and going straight, respectively. And $\varepsilon$ represents the judgment threshold, with a default setting of 25.

\begin{figure*}[t]
    \centering
    \includegraphics[width=1.0\linewidth]{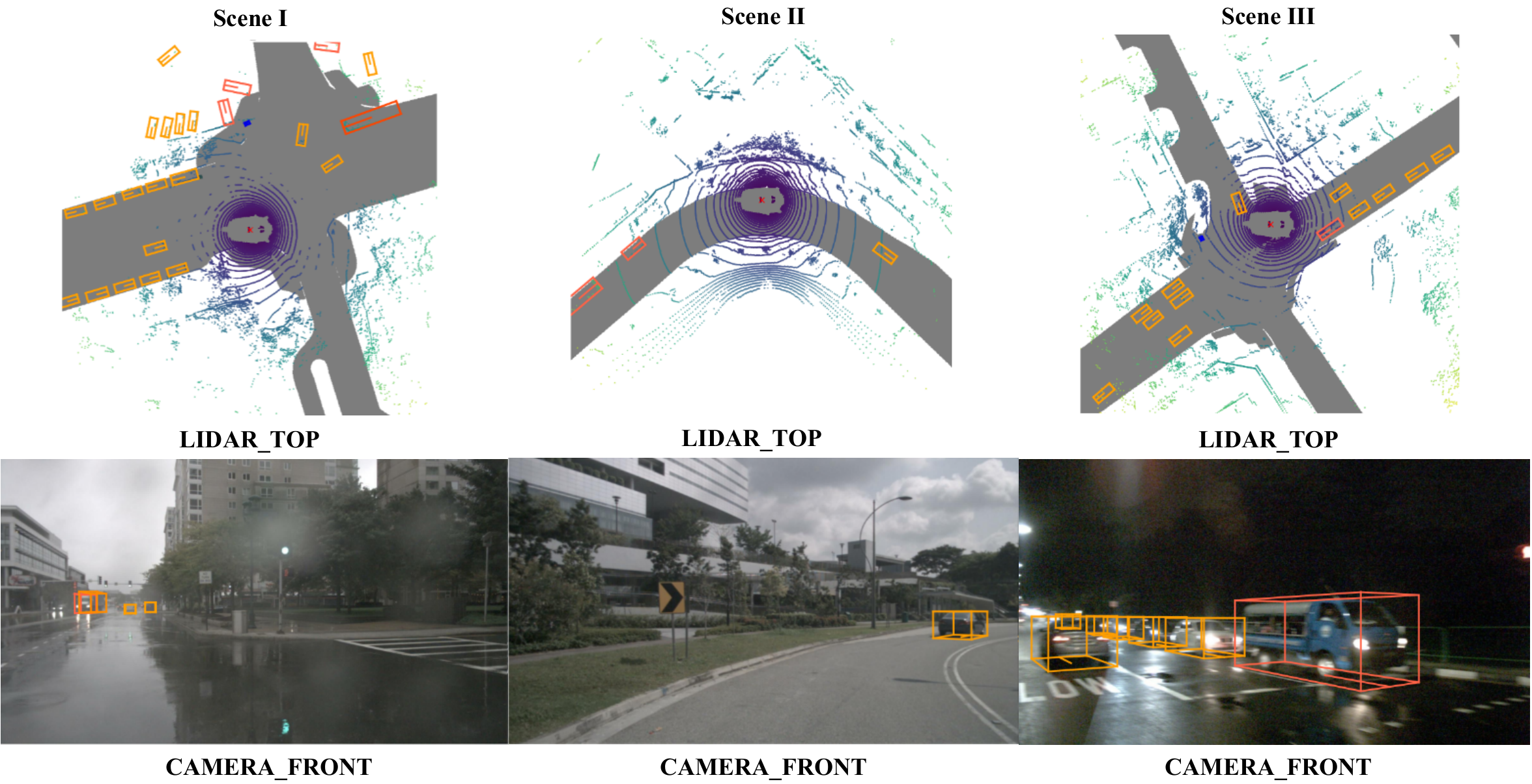}
    \caption{Visualization of turning scenarios in the \textbf{Turning-nuScenes dataset}. ``LIDAR\_TOP'' represents the visualization of the corresponding scene from BEV. While ``CAMERA\_FRONT'' refers to the images captured by the front camera of the ego vehicle in the respective scene.}
    \label{fig:vis_turn_scene}
\end{figure*}

\noindent 3) \textbf{Generate Dataset.} After sample select, we obtained a series of sample\_tokens associated with turning scenarios, denoted as $sample\_token_{select}$. Based on the mapping relationship $dict_{sa}^{sc}$ from \textit{scene\_token} to \textit{sample\_token}, we derive a series of driving scenarios involving the ego vehicle's turning maneuvers. The Turning-NuScene dataset comprises 17 scenes with 680 samples and includes diverse urban turning scenarios, such as intersections, T-junctions, roundabouts, traffic islands, and alleyway turns. The visualization of some data from Turning-nuScenes dataset is shown in Fig. \ref{fig:vis_turn_scene}.

% 首先，我们根据NuS数据集的标定，提取出验证集的全部数据信息。Next,对于验证集中每一个sample数据，我们提取其中自车未来一段时间内的轨迹。考虑到在行驶过程中，驾驶行进方向为世界坐标系下的y轴方向，当汽车出现转弯时，车辆的x坐标会发生显著的变化。因此，我们根据自我车辆x坐标的位置变化来判断车辆在未来是否存在转弯动作并记录下该sample数据的唯一标识符号{sample_token}。最后根据场景唯一标识符和数据标识符的映射关系，我们得到系列包含转弯情形下的驾驶场景数据并构成t-nus数据集。t——nus数据集有1079个sample数据构成，并涵盖多种城市转弯场景，包括十字路口，T字路口，环形交叉口，交通岛以及小巷口转弯等场景。部分场景的可视化如图一所示
%Specifically, for each sample in the nuScenes dataset, we extract the trajectory data of the ego vehicle for the next 3 seconds. Since the vehicle's heading direction aligns with the y-axis during driving, we determine whether the vehicle will make a turn in the future based on the changes in its x-coordinate position. If a turning action is detected, we save the scene token associated with that sample. Ultimately, we selected data from \textbf{19} turning scenarios to construct the Turning-nuScenes dataset. To facilitate reproducibility for other researchers, we include the inference code in Listing \ref{fig:code}.

\begin{table*}[t]
\small
\centering
  \caption{Planning results on the $\operatorname{Turning\text{-}nuScenes}$ validation dataset. $\operatorname{UniAD}$~\cite{uniad} and $\operatorname{VAD}$~\cite{jiang2023vad} are SOTA end-to-end deterministic planning methods. $\operatorname{SparseDrive}$~\cite{sun2024sparsedrive} is a SOTA end-to-end multi-modal trajectory planning method. \textcolor{red}{Red} indicates improvement. We follow the ST-P3~\cite{ST_P3} evaluation metric.}\vspace{-2ex}
  \renewcommand\arraystretch{1.0}
  \setlength{\tabcolsep}{1.5mm} % Adjust column spacing
  \resizebox{\linewidth}{!}{
  \begin{tabular}{lcccc cccc cccc}
\toprule
\multirow{2}{*}{$\operatorname{Method}$} & \multicolumn{4}{c}{$\operatorname{L2\ (m)}\downarrow$} & \multicolumn{4}{c}{$\operatorname{Col.\ Rate\ (\%)}\downarrow$} & \multicolumn{4}{c}{$\operatorname{TPC\ (m)}\downarrow$} \\
\cmidrule(lr){2-5} \cmidrule(lr){6-9} \cmidrule(lr){10-13}
& 1s & 2s & 3s & $\operatorname{Avg.}$ & 1s & 2s & 3s & $\operatorname{Avg.}$ & 1s & 2s & 3s & $\operatorname{Avg.}$ \\
\midrule
$\operatorname{UniAD}$~\cite{uniad} & 0.52 & 0.88 & 1.64 & \cellcolor{gray!15}1.01 & 0.16 & 0.51 & 1.41 & \cellcolor{gray!15}0.69 & 0.47 & 0.81 & 1.58 & \cellcolor{gray!15}0.95 \\
$\operatorname{VAD}$~\cite{jiang2023vad} & 0.48 & 0.80 & 1.55 & \cellcolor{gray!15}0.94 & 0.07 & 0.41 & 1.20 & \cellcolor{gray!15}0.56 & 0.38 & 0.78 & 1.51 & \cellcolor{gray!15}0.89 \\
$\operatorname{SparseDrive}$~\cite{sun2024sparsedrive} & 0.35 & 0.77 & 1.46 & \cellcolor{gray!15}0.86 & 0.04 & 0.17 & 0.98 & \cellcolor{gray!15}0.40 & 0.34 & 0.70 & 1.33 & \cellcolor{gray!15}0.79 \\

\rowcolor{gray!15} $\operatorname{MomAD\ (Ours)}$ & 
\textbf{0.33}\textit{\fontsize{6}{0}\selectfont\textcolor{red}{\textbf{-0.02}}} & 
\textbf{0.70}\textit{\fontsize{6}{0}\selectfont\textcolor{red}{\textbf{-0.07}}} & 
\textbf{1.24}\textit{\fontsize{6}{0}\selectfont\textcolor{red}{\textbf{-0.22}}} & 
\textbf{0.76}\textit{\fontsize{6}{0}\selectfont\textcolor{red}{\textbf{-0.10}}} & 
0.03\textit{\fontsize{6}{0}\selectfont\textcolor{red}{\textbf{-0.01}}} & 
\textbf{0.13}\textit{\fontsize{6}{0}\selectfont\textcolor{red}{\textbf{-0.04}}} & 
\textbf{0.79}\textit{\fontsize{6}{0}\selectfont\textcolor{red}{\textbf{-0.19}}} & 
\textbf{0.32}\textit{\fontsize{6}{0}\selectfont\textcolor{red}{\textbf{-0.08}}} & 
0.32\textit{\fontsize{6}{0}\selectfont\textcolor{red}{\textbf{-0.02}}} & 
\textbf{0.54}\textit{\fontsize{6}{0}\selectfont\textcolor{red}{\textbf{-0.16}}} & 
\textbf{1.05}\textit{\fontsize{6}{0}\selectfont\textcolor{red}{\textbf{-0.28}}} & 
\textbf{0.63}\textit{\fontsize{6}{0}\selectfont\textcolor{red}{\textbf{-0.16}}} \\
\bottomrule
\end{tabular}}
\label{tab_nuscenes_planning_turning_appendix}
\end{table*}

\subsection{Implementation Details}
\label{sec:Implementation_Details}
\noindent The training process of MomAD is divided into two stages following SparseDrive~\cite{sun2024sparsedrive}. In stage 1, we train the sparse perception module, including 3D object detection, multi-object tracking, and online mapping, from scratch to learn sparse scene representations. In stage 2, we train the sparse perception, motion, and planning modules without freezing the weights of the sparse perception module. For MomAD, we use ResNet50~\cite{resnet} as backbone network and the input image size is 256 $\times$ 704. For detection, the perception range is a circle with a radius of 55m. For online mapping, the perception range is 60m $\times$ 30m longitudinally and laterally. For motion and planning, the number of stored frames $H$ in the instance memory queue is set to 3, and the number of modes $K_m$ in motion is set to $6$, accounting for six trajectory proposals.
All experiments are conducted on 8 NVIDIA RTX 4090 24GB GPUs. 

\noindent \textbf{Stage-1 Overall Objectives.} In alignment with SparseDrive~\cite{sun2024sparsedrive} and VAD~\cite{jiang2023vad}, MomAD does not enforce tracking constraints during the identity assignment process. As a result, we do not include a tracking loss in our framework. The loss function for the supervised process during the first phase is defined as follows,
\begin{equation}
    \label{eq:loss_stage1}
    \begin{aligned}
\mathbf{L}_{1} = \mathbf{L_D}  + \mathbf{L_M}.
    \end{aligned}
\end{equation}

\noindent \textbf{Stage-2 Overall Objectives.} MomAD is trained utilizing the losses from all tasks, which include 3D object detection, multi-object tracking, online mapping, motion prediction, and planning. This training is conducted over a duration of 10 epochs, employing a total batch size of 48 and a learning rate of  $3\times e^{-4}$. The loss function for the supervised process during this stage is defined as follows,
\begin{equation}
    \label{eq:loss_stage2}
    \begin{aligned}
\mathbf{L}_{2} = \mathbf{L_D} + \mathbf{L_M} + \mathbf{L_{MP}}.
    \end{aligned}
\end{equation}

\noindent \textbf{Detection Loss.} The detection loss is formulated as a linear combination of the Focal Loss~\cite{focalloss} for classification and the L1 Loss for box regression.
\begin{equation}
    \label{eq:loss_det}
    \begin{aligned}
\mathbf{L_D} = \lambda_c \mathbf{L_D}_c + \lambda_r \mathbf{L_D}_r,
    \end{aligned}
\end{equation}
which $\lambda_c$ and $\lambda_r$ are set to 2 and 0.25, respectively.

\noindent \textbf{Online Mapping Loss.}
In accordance with VAD~\cite{jiang2023vad} and SparseDrive~\cite{sun2024sparsedrive}, we define the online mapping loss as the following equation,
\begin{equation}
    \label{eq:loss_map}
    \begin{aligned}
\mathbf{L_M} =  \lambda_c \mathbf{L_M}_c + \lambda_r \mathbf{L_M}_r,
    \end{aligned}
\end{equation}
which $\lambda_c$ and $\lambda_r$ are set to 1 and 10, respectively.

\noindent\textbf{Motion and Planning Loss.} We compute the average displacement error (ADE) between the multi-modal outputs and the ground truth trajectory. The trajectory with the lowest ADE is designated as the positive sample, while the remaining trajectories are treated as negative samples. In addition, for the planning component, the ego state is also predicted. We employ Focal Loss for classification and L1 Loss for regression,
\begin{equation}
    \label{eq:loss_motion}
    \begin{aligned}
\mathbf{L_{MP}} = 
\lambda^{m}_c \mathbf{L_{MO}}_c +  
\lambda^{m}_r \mathbf{L_{MO}}_r + 
\lambda^{p}_c \mathbf{L_{p}}_c + 
\lambda^{p}_r \mathbf{L_{p}}_r +  
\lambda^{p}_s \mathbf{L_{s}}
    \end{aligned}
\end{equation}
which $\lambda^{m}_c$ and $\lambda^{m}_r$ are set to 0.2 and 0.2, $\lambda^{p}_c$, $\lambda^{p}_r$ and  $\lambda^{p}_s$ are set to 0.5, 1.0 and 1.0, respectively.

\begin{table}[t]
\Large
\centering
  \caption{Long trajectory planning results on the $\operatorname{nuScenes}$ and $\operatorname{Turning\text{-}nuScenes}$ validation sets. We train models for 10 epochs for 6s-horizon prediction. $\operatorname{T\text{-}nuScenes}$ indicates the challenging $\operatorname{Turning\text{-}nuScenes}$. We follow the ST-P3~\cite{ST_P3} evaluation metric.}\vspace{-1ex}
  \renewcommand\arraystretch{1.0}
  \setlength{\tabcolsep}{1.0mm}
  \resizebox{\linewidth}{!}{
  \begin{tabular}{ll ccc ccc ccc}
\toprule
\multirow{2}{*}{$\operatorname{Split}$} & \multirow{2}{*}{$\operatorname{Method}$} & \multicolumn{3}{c}{$\operatorname{L2\ (m)}\downarrow$} & \multicolumn{3}{c}{$\operatorname{Col.\ Rate\ (\%)}\downarrow$} & \multicolumn{3}{c}{$\operatorname{TPC\ (m)}\downarrow$} \\
\cmidrule(lr){3-5} \cmidrule(lr){6-8} \cmidrule(lr){9-11}
& & 4s & 5s & 6s & 4s & 5s & 6s & 4s & 5s & 6s \\
\midrule
\multirow{5}{*}{$\operatorname{nuScenes}$} 
& $\operatorname{UniAD}$~\cite{uniad} & 1.91 & 2.57 & 3.21 & 0.91 & 1.66 & 2.51 & 1.49 & 1.81 & 2.41 \\
& $\operatorname{VAD}$~\cite{jiang2023vad} & 1.82 & 2.23 & 3.01 & 0.89 & 1.71 & 2.41 & 1.55 & 1.73 & 2.17 \\
& $\operatorname{SparseDrive}$~\cite{sun2024sparsedrive} & 1.75 & 2.32 & 2.95 & 0.87 & 1.54 & 2.33 & 1.33 & 1.66 & 1.99 \\
& \cellcolor{gray!15} $\operatorname{MomAD}$ & 
\cellcolor{gray!15}1.67 & \cellcolor{gray!15}1.98 & \cellcolor{gray!15}2.45 & 
\cellcolor{gray!15}0.83 & \cellcolor{gray!15}1.43 & \cellcolor{gray!15}2.13 & 
\cellcolor{gray!15}1.19 & \cellcolor{gray!15}1.45 & \cellcolor{gray!15}1.61 \\
& & 
\textit{\textcolor{blue}{\textbf{-0.09}}} & 
\textit{\textcolor{blue}{\textbf{-0.34}}} & 
\textit{\textcolor{blue}{\textbf{-0.50}}} & 
\textit{\textcolor{blue}{\textbf{-0.04}}} & 
\textit{\textcolor{blue}{\textbf{-0.11}}} & 
\textit{\textcolor{blue}{\textbf{-0.20}}} & 
\textit{\textcolor{blue}{\textbf{-0.14}}} & 
\textit{\textcolor{blue}{\textbf{-0.21}}} & 
\textit{\textcolor{blue}{\textbf{-0.38}}} \\
\midrule
\multirow{5}{*}{$\operatorname{T\text{-}nuScenes}$} 
& $\operatorname{UniAD}$~\cite{uniad} & 2.45 & 2.98 & 3.76 & 1.21 & 1.99 & 3.25 & 1.81 & 2.75 & 3.42 \\
& $\operatorname{VAD}$~\cite{jiang2023vad} & 2.27 & 2.87 & 3.46 & 1.08 & 1.86 & 2.81 & 1.68 & 2.56 & 3.21 \\
& $\operatorname{SparseDrive}$~\cite{sun2024sparsedrive} & 2.07 & 2.71 & 3.36 & 0.91 & 1.71 & 2.57 & 1.54 & 2.31 & 2.90 \\
& \cellcolor{gray!15} $\operatorname{MomAD}$ & 
\cellcolor{gray!15}1.80 & \cellcolor{gray!15}2.07 & \cellcolor{gray!15}2.51 & 
\cellcolor{gray!15}0.85 & \cellcolor{gray!15}1.57 & \cellcolor{gray!15}2.31 & 
\cellcolor{gray!15}1.37 & \cellcolor{gray!15}1.58 & \cellcolor{gray!15}1.93 
\\
& & 
\textit{\textcolor{red}{\textbf{-0.27}}} & 
\textit{\textcolor{red}{\textbf{-0.64}}} & 
\textit{\selectfont\textcolor{red}{\textbf{-0.85}}} & 
\textit{\selectfont\textcolor{red}{\textbf{-0.06}}} & 
\textit{\selectfont\textcolor{red}{\textbf{-0.14}}} & 
\textit{\textcolor{red}{\textbf{-0.26}}} & 
\textit{\textcolor{red}{\textbf{-0.17}}} & 
\textit{\textcolor{red}{\textbf{-0.73}}} & 
\textit{\textcolor{red}{\textbf{-0.97}}} \\
\bottomrule
\end{tabular} }
\label{tab_nuscenes_Turning_nuscenes_planning_6s_appendix}
\end{table}

\begin{table}[!htp]
% \tiny
\centering\vspace{-2ex}
\addtolength{\tabcolsep}{0.1pt}
\caption{Robustness analysis on $\operatorname{nuScenes-C}$ \cite{zhujun_benchmarking}.}
     \vspace{-0.8em}
  \renewcommand\arraystretch{0.9}
  \tabcolsep=0.3mm %%%%%%%%%
  \resizebox{\linewidth}{!}{
  %\begin{tabular*}{\linewidth} {@{}@{\extracolsep{\fill}}!{\color{white}\vline}l|c|c|c|c|c|c|c|c|c|c|c|c @{}}
  \begin{tabular}{ll cccc cccc}
    \toprule
\multirow{2}{*}{\textbf{$\operatorname{Scene}$}}&   \multirow{2}{*}{\textbf{$\operatorname{Method}$}}& \multicolumn{2}{c}{\textbf{$\operatorname{Detection}$}}     & $\operatorname{Tracking}$ & $\operatorname{Mapping}$ & $\operatorname{Motion}$  & \multicolumn{3}{c}{\textbf{$\operatorname{Planning}$}}   \\
   \cmidrule(lr){3-4} \cmidrule(lr){5-5} \cmidrule(lr){6-6} \cmidrule(lr){7-7} \cmidrule(lr){8-10}
 &  &$\operatorname{mAP}\uparrow$&$\operatorname{NDS}\uparrow$ &     
   $\operatorname{AMOTA}\uparrow$ 
   & $\operatorname{mAP}\uparrow$  
   & $\operatorname{mADE}\downarrow$ 
   & $\operatorname{L2}\downarrow$& $\operatorname{Col.}\downarrow$  & $\operatorname{TPC}\downarrow$
   \\
\midrule

\multirow{2}{*}{$\operatorname{Clean}$}&$\operatorname{SparseDrive}$ &0.418& 0.525& 0.386& 55.1& 0.62&0.61&0.08&0.57 \\
&\cellcolor{gray!15}$\operatorname{MomAD}$   &\cellcolor{gray!15} \textbf{0.423}&\cellcolor{gray!15} \textbf{0.531}&\cellcolor{gray!15} \textbf{0.391}& \cellcolor{gray!15}\textbf{55.9}&\cellcolor{gray!15}\textbf{0.61}&\cellcolor{gray!15}\textbf{0.60}&\cellcolor{gray!15}\textbf{0.09}&\cellcolor{gray!15}\textbf{0.54} \\
\midrule 
\multirow{2}{*}{$\operatorname{Snow}$}&$\operatorname{SparseDrive}$ &0.140& 0.161& 0.133& 22.3& 0.95& 0.85& 0.30& 0.79 \\
&\cellcolor{gray!15}$\operatorname{MomAD}$   &\cellcolor{gray!15}\textbf{0.172} &\cellcolor{gray!15}\textbf{0.195}& \cellcolor{gray!15}\textbf{0.169}& \cellcolor{gray!15}\textbf{27.9}& \cellcolor{gray!15}\textbf{0.72}& \cellcolor{gray!15}\textbf{0.71}& \cellcolor{gray!15}\textbf{0.18} &\cellcolor{gray!15}\textbf{0.66} \\
\midrule
\multirow{2}{*}{$\operatorname{Rain}$}&$\operatorname{SparseDrive}$ & 0.232&  0.254&  0.198 & 30.7&  0.96&  0.87&  0.31&  0.83 \\
&\cellcolor{gray!15}$\operatorname{MomAD}$   &\cellcolor{gray!15}\textbf{0.270}& \cellcolor{gray!15}\textbf{0.293} & \cellcolor{gray!15}\textbf{0.222} & \cellcolor{gray!15}\textbf{34.8} & \cellcolor{gray!15}\textbf{0.71} & \cellcolor{gray!15}\textbf{0.67} & \cellcolor{gray!15}\textbf{0.18} & \cellcolor{gray!15}\textbf{0.67}  \\
\midrule
\multirow{2}{*}{$\operatorname{Fog}$}&$\operatorname{SparseDrive}$ &0.294& 0.312&0.260& 41.2& 0.93& 0.84& 0.36& 0.80 \\
&\cellcolor{gray!15}$\operatorname{MomAD}$   & \cellcolor{gray!15}\textbf{0.348} & \cellcolor{gray!15}\textbf{0.356} & \cellcolor{gray!15}\textbf{0.299} & \cellcolor{gray!15}\textbf{43.2} & \cellcolor{gray!15}\textbf{0.68} & \cellcolor{gray!15}\textbf{0.64} & \cellcolor{gray!15}\textbf{0.19} & \cellcolor{gray!15}\textbf{0.61}  \\

\bottomrule
\end{tabular} }
\label{tab_robust_appendix}
\end{table}

\begin{figure*}[t]
    \centering
    \includegraphics[width=1.0\linewidth]{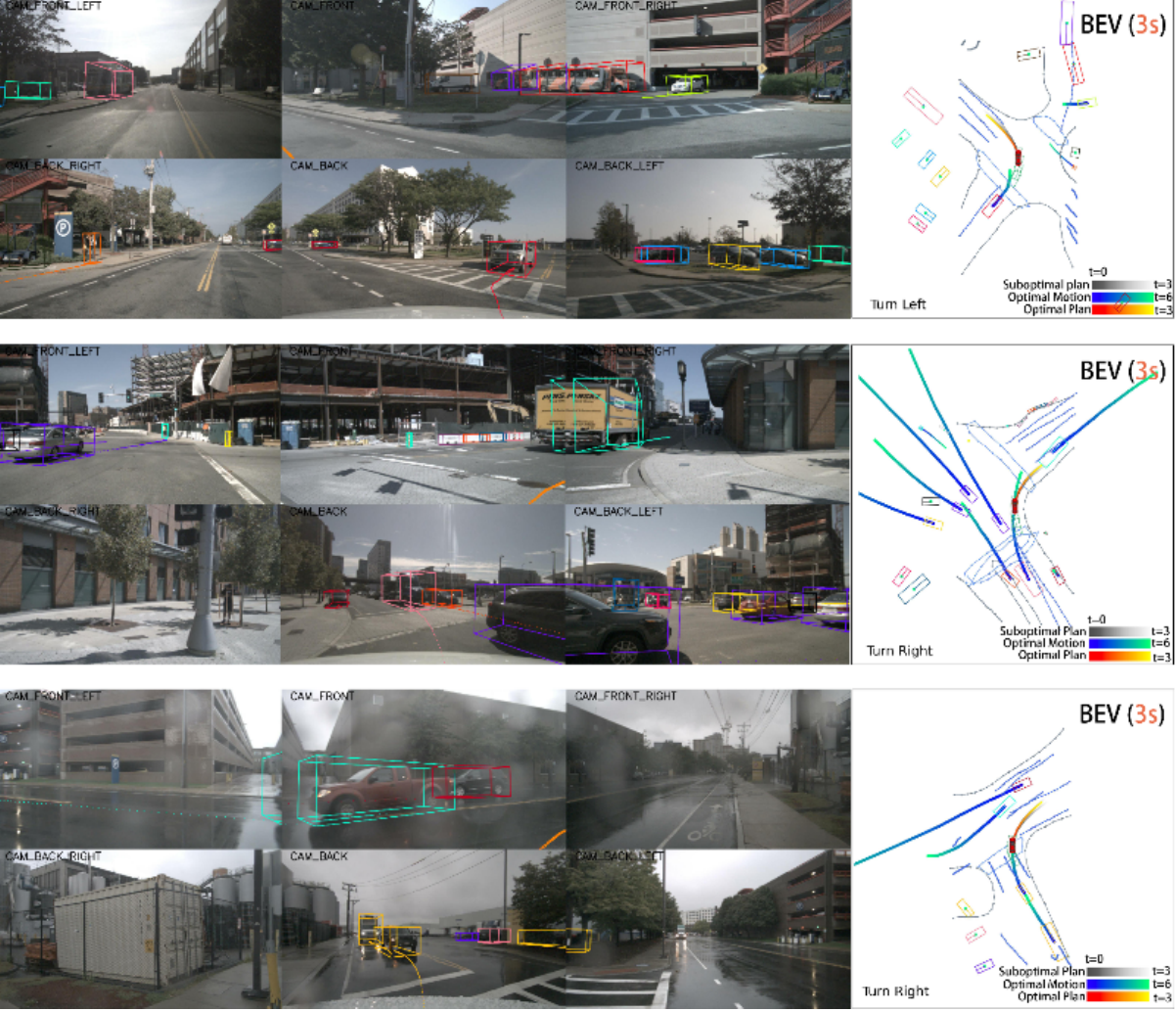}
    \caption{Visualization results (Planning for \textbf{3s} Trajectory Prediction). We visualize results for detection, online mapping, motion prediction, and planning. MomAD demonstrates stable and temporally consistent planning across various complex turning scenarios, especially in crowded environments. For motion prediction, we present the model's selected trajectory from multi-modal proposals, with each trajectory spanning a 6-second duration. For planning, the selected (optimal) trajectory is visualized in \textcolor{red}{\textbf{red}}, alongside two suboptimal (proposal) multi-modal trajectories in \textcolor{gray}{\textbf{gray}}.}
    \label{fig:turn_3s}
\end{figure*}

\begin{figure*}[t]
    \centering
    \includegraphics[width=1.0\linewidth]{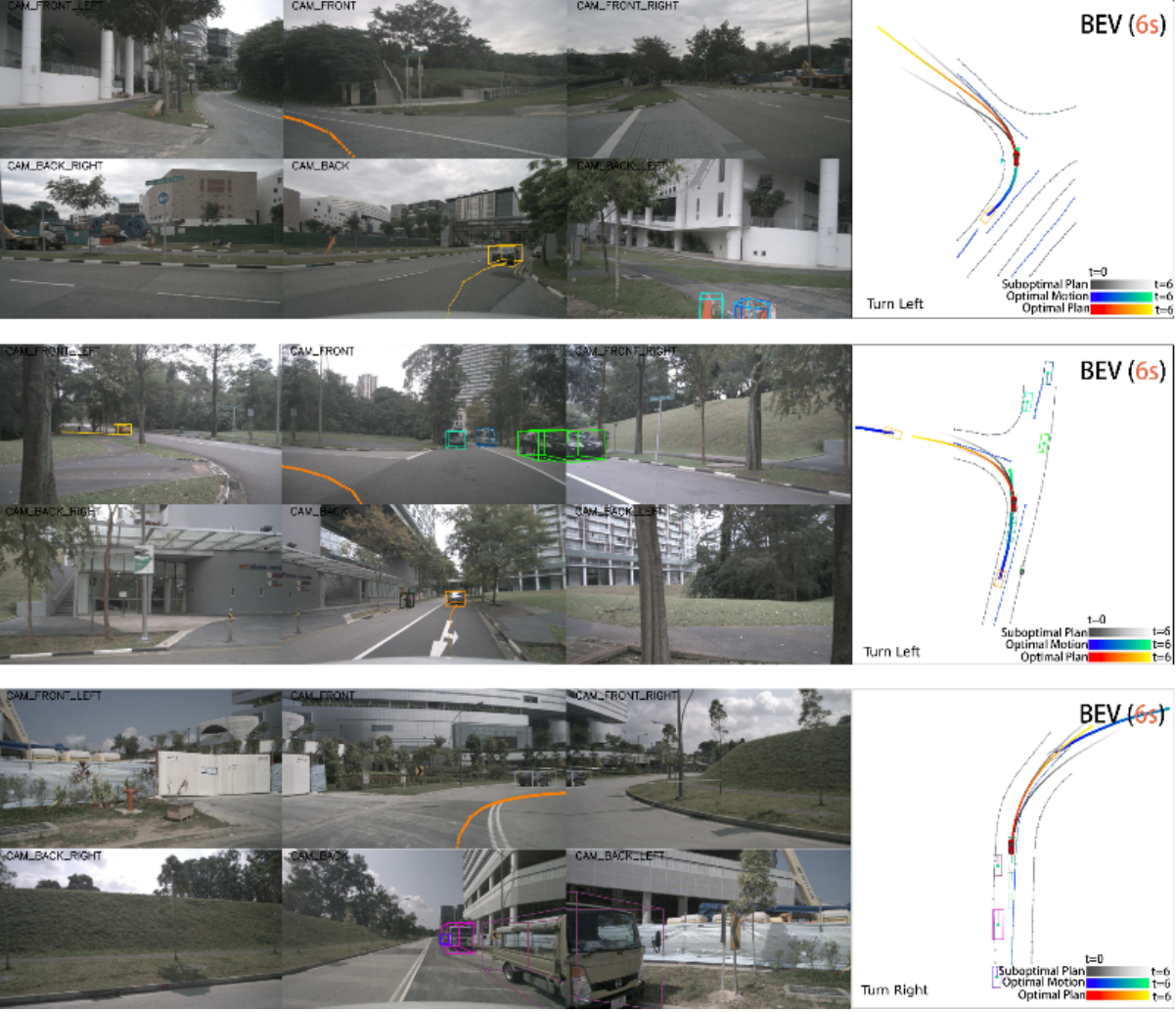}
    \caption{Visualization results (Planning for \textbf{6s} Trajectory Prediction). Long-horizon trajectories often face greater temporal consistency issues. We present 6-second trajectory prediction results to demonstrate how MomAD addresses these inconsistencies. Despite the increased challenge of long-horizon trajectories, MomAD continues to exhibit robust and stable performance. For motion prediction, we show the trajectory with the highest score from the model’s output, each spanning 6 seconds. For planning, the selected (optimal) trajectory is visualized in \textcolor{red}{\textbf{red}}, accompanied by two suboptimal (proposal) multi-modal trajectories in \textcolor{gray}{\textbf{gray}}.}
    \label{fig:turn_6s}
\end{figure*}

\begin{figure*}[t]
    \centering
    \includegraphics[width=1.0\linewidth]{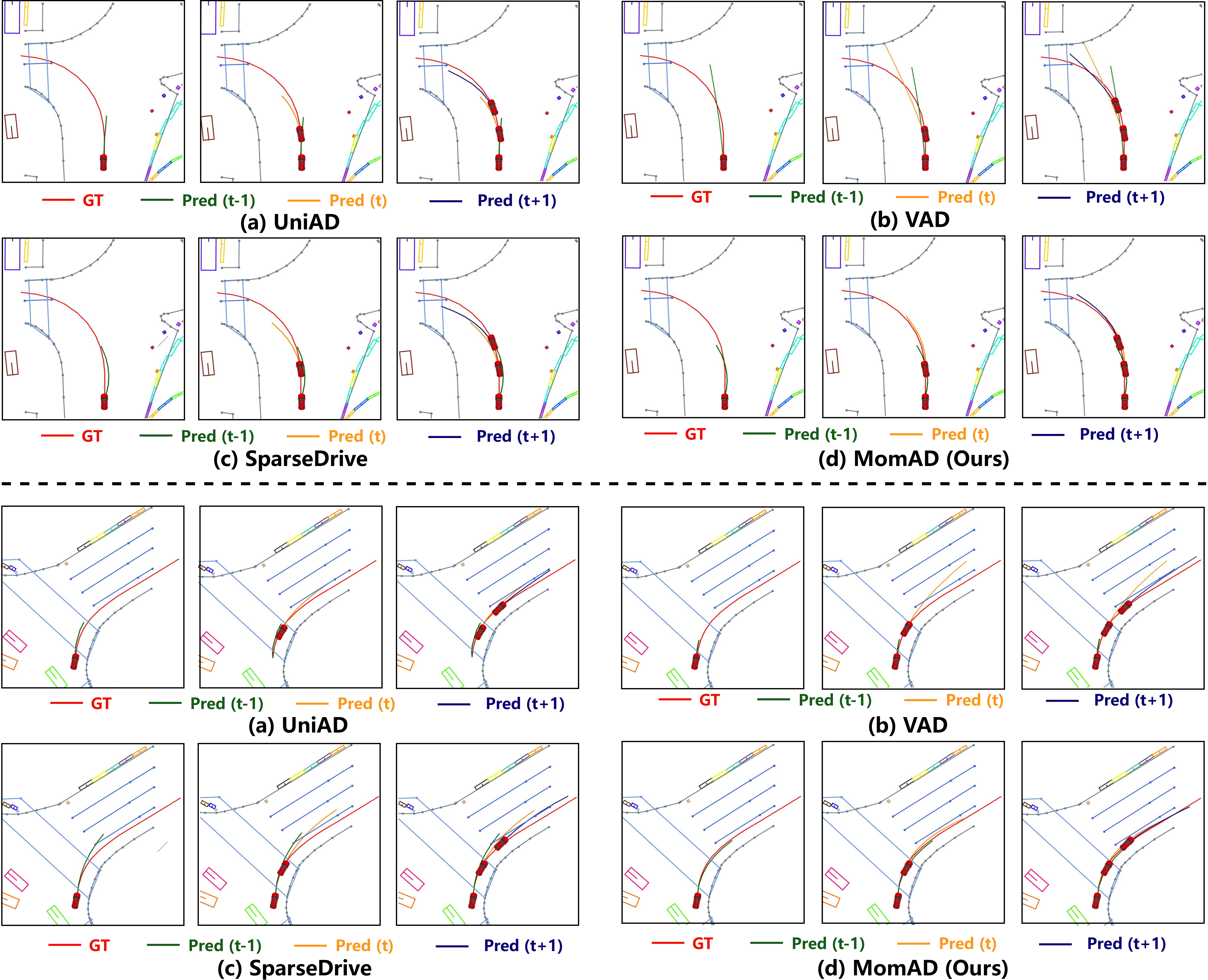}
    \caption{More visualization results of MomAD with SOTA methods across multiple frames.}
    \label{fig:vis_adx}
\end{figure*}

\subsection{More Planning Results}
\label{sec:Results}
We have extended the results of Tables 2 and 3 in the main by including UniAD~\cite{uniad} and VAD~\cite{jiang2023vad} to provide additional experimental data. As shown in Tables \ref{tab_nuscenes_planning_turning_appendix} and \ref{tab_nuscenes_Turning_nuscenes_planning_6s_appendix}, our conclusion is consistent with those presented in the main text: end-to-end autonomous driving methods represented by UniAD~\cite{uniad}, VAD~\cite{jiang2023vad}, and SparseDrive~\cite{sun2024sparsedrive} suffer challenges in turning scenarios. Our TPC metric demonstrates issues of robustness in temporal consistency, as these methods enable seamless integration of perception and planning but often rely on one-shot trajectory prediction, which may lead to unstable control and vulnerability to occlusions in single-frame perception.
Overall, our proposed MomAD addresses key challenges in planning stability and robustness for end-to-end autonomous driving systems.

\subsection{Detailed Result Analysis on Robustness}
\label{sec:Robustness}
As shown in Table \ref{tab_robust_appendix}, we furthur evaluated MomAD on \textbf{nuScenes-C} \cite{zhujun_benchmarking}, which benchmarks robustness against diverse corruptions including extreme weathers. Our MomAD consistently outperforms SparseDrive across all tasks, by \textbf{22.9\%} (\textit{detection}), \textbf{27.1\%} (\textit{tracking}), \textbf{25.1\%} (\textit{mapping}), \textbf{24.2\%} (\textit{motion}), and \textbf{40.0\%} (planning) on average. These results highlight the robustness of MomAD against various noise perturbations.

\subsection{More Qualitative Study of Planning Results}
\label{sec:Qualitative}
To better illustrate the exceptional planning capabilities of MomAD, we selected planning results from complex traffic scenarios for visualization, such as turning maneuvers and congested scenes. We provide three qualitative results: (1) planning for 3s trajectory prediction, (2) planning for 6 trajectory prediction, and (3) trajectory prediction across multiple frames.

\noindent \textbf{(1) Planning for 3s Trajectory Prediction.} Consistent with most end-to-end autonomous driving methods, we provide conventional 3-second prediction results, including the selected optimal trajectory and multi-modal proposal trajectory, as well as the optimal motion trajectory. As shown in Fig. \ref{fig:turn_3s}, MomAD performs well across various turning scenarios, successfully executing large-angle turns without any collisions. 

\noindent \textbf{(2) Planning for 6s Trajectory Prediction.} Unlike most end-to-end autonomous driving methods, we offer long-horizon trajectory predictions with a 6-second horizon. As depicted in Fig. \ref{fig:turn_6s}, even under more challenging conditions, MomAD maintains superior planning performance. Specifically, the predicted trajectory remains smooth and consistent even over a long-horizon trajectory. This strong performance can be attributed to the proposed MomAD's effective use of historical trajectory data. By incorporating past trajectories, MomAD is able to predict and adapt to dynamic changes in the environment, ensuring smoother navigation and more accurate decision-making during turns.

\noindent \textbf{(3) Trajectory Prediction across Multiple Frames.} 
As shown in Figure \ref{fig:vis_adx}, we present two multi-frame qualitative results to highlight the consistency and robustness of the proposed MomAD method. In the turning scenario, MomAD generates a smooth and accurate trajectory, demonstrating its ability to avoid oscillatory behavior during the planning process—a critical factor for ensuring driving safety. In conclusion, the visual results clearly illustrate the superior performance of MomAD in trajectory planning.

\end{document}